\documentclass[preprint]{elsarticle}

\usepackage{palatino,epsfig,latexsym}
\usepackage{amsmath} 
\usepackage{amssymb} 
\usepackage{color} 
\usepackage{enumitem} 
\usepackage{xcolor}
\usepackage{graphicx}

\usepackage{listings}
\usepackage{url}
\usepackage[pdfpagelabels,bookmarks,hyperindex,hyperfigures]{hyperref}

\journal{European Journal of Operational Research}

\usepackage{xcolor}
\usepackage[normalem]{ulem}

\newcommand{\added}[1]{#1}
\newcommand{\removed}[1]{}
\newcommand{\replaced}[2]{#2}
\newcommand{\msgtoreviewer}[1]{}{}

\begin{document}

\clearpage

\begin{frontmatter}

\title{Metaheuristics ``In the Large''}

\author{Jerry Swan\corref{cor1}}
\author{Steven Adriaensen}
\author{Alexander E.\ I.\ Brownlee}
\author{Kevin Hammond}
\author{\\Colin G.\ Johnson}
\author{Ahmed Kheiri}
\author{Faustyna Krawiec}
\author{J.\ J.\ Merelo}
\author{\\Leandro L.\ Minku}
\author{Ender \"Ozcan}
\author{Gisele L.\ Pappa}
\author{Pablo Garc\'ia-S\'anchez}
\author{\\Kenneth S\"orensen}
\author{Stefan Vo{\ss}}
\author{Markus Wagner}
\author{David R.\ White}


\cortext[cor1]{
Corresponding author: jerry.swan@york.ac.uk. University of York, York YO10 5DD, UK.
}


\hyphenation{principled}

\begin{abstract}
\msgtoreviewer{In this version of the article, we have tried to highlight the changes we made in the latest revision (\removed{removed text}, \added{added text}).}

Following  decades of  sustained improvement, metaheuristics are  one  of the great success stories of optimization research. However, in order for research in metaheuristics to avoid fragmentation and a lack of reproducibility, there is a pressing need for stronger scientific and computational infrastructure to support the development, analysis and comparison of new approaches. \added{To this end, we present the vision and progress of the ``Metaheuristics `In the Large' '' project. The conceptual uderpinnings of the project are: truly extensible algorithm templates that support reuse without modification, white box problem descriptions that provide generic support for the injection of domain specific knowledge, and remotely accessible frameworks, components and problems that will enhance reproducibility and accelerate the field's progress.} We argue that, via \removed{such} principled choice of infrastructure support, the field can pursue a higher level of scientific enquiry.  We describe our vision and report on progress, showing how the adoption of common protocols for all metaheuristics can help liberate the potential of the field, easing the exploration of the design space of metaheuristics.

\end{abstract}

\end{frontmatter}



\section{Introduction}
Optimization problems have myriad real world applications \cite{HoosStutzle:04} and have motivated a wealth of research since before the advent of the digital computer \cite{Dantzig:1990:OSM}. Recent decades have seen enormous progress in the discipline of metaheuristic optimization\footnote{In this article (and in the spirit of S\"orensen and Glover, \cite{SorensenGlover2013}), we reserve the term ``metaheuristic' for the generic, cross domain framework and the term ``heuristic'' for a customization of such a framework to one or more specific domains.}. In contrast to \emph{exact} approaches that guarantee optimality, a metaheuristic is an iterative master process that guides and modifies the operations of subordinate heuristics to efficiently produce high-quality solutions. At each iteration, it manipulates either a complete (or partial) single solution or else a collection of such solutions. The subordinate heuristics may be high or low level procedures, or a simple local search, or just a construction method. The family of metaheuristics includes, but is not limited to: adaptive memory procedures, tabu search, swarm intelligence, greedy randomized adaptive search, variable neighborhood search, evolutionary methods, genetic algorithms, scatter search, neural networks, simulated annealing, and their hybrids \cite{SorensenGlover2013,PELLERIN2020395}. 
One of the major advantages of metaheuristics is that they are abstract search methods \cite{DecomposingMetaheuristicOperations}: the underlying search logic can be applied to any problem which can be decomposed into a few elementary aspects, namely solution representation, solution quality evaluation and some notion of locality. The latter denotes the ability to generate neighboring solutions via a heuristically-informed function of one or more incumbent solutions.

Very broadly speaking, one might distinguish between \emph{classical} OR 
and metaheuristic approaches with respect \replaced{the}{to} the former's emphasis on \emph{analytic} methods and the latter's emphasis on \emph{empirical} ones \cite{recharacterizationhh}. An analytic approach uses problem domain information (and typically also \emph{a priori} human ingenuity) to derive effective algorithms for search components --- edge-assembly (EAX) crossover for the Travelling Salesman Problem being one such example~\cite{DBLP:conf/icga/NagataK97}. \added{Indeed, the pre-eminent successes of OR often arise directly from a direct match between the solver (e.g.\ linear programming) and the analytic characteristics of the problem. In contrast, the use of analytic problem characteristics to choose solvers is not part of mainstream metaheuristics.} The empirical approach performs configuration tuning (by hand, using statistical design or some Machine Learning technique) to create a metaheuristic biased either offline by a target distribution of problem instances and/or online by the search trajectory. 

Despite the significant progress in metaheuristic optimization research, it is increasingly acknowledged within the scientific community that the field suffers from a duplication of effort and siloing of research between papers, groups, and software frameworks. This lack of re-use is evident at both  \emph{conceptual} and \emph{implementation} levels:
~ \\  
\begin{itemize}
    \item Conceptual: an over-reliance on reasoning by metaphor \cite{Sor13} \removed{\cite{BIRGIN2015421}} hides commonalities between algorithms, leading to the repeated discovery of the same ideas and heuristics, and widespread duplication of research effort.
    \item Implementation: despite past efforts in developing software frameworks, there is a tendency to re-implement metaheuristics from scratch, hindering reproducibility and replicability \cite{micstandards2015,Pamp1512:Generic,AOCPECJ}.     
\end{itemize}
~ \\  
This duplication of effort limits scientific progress; instead of building a cohesive body of knowledge consisting of robust scientific conclusions, accumulated wisdom in the field is more akin to ``folklore'': observations over individual algorithms and optimization problems, without a structured underlying narrative. In the words of Pampar\`{a}, there is a ``throw away'' research culture \cite{Pamp1512:Generic}: it is difficult to locate and compare against prior-art, and there is a lack of understanding as to which heuristics work best and in what context.  Despite the availability of many software libraries, it is difficult to reuse existing implementations --- in particular, it is extremely difficult to combine heuristics from different libraries or incorporate domain-specific knowledge in a general manner, which has hindered the creation of easily testable and deployable metaheuristic pipelines.

While many authors have identified and critiqued a lack of rigor and weak empirical method in the field (e.g. \cite{Johnson2002, Kendall2016, collberg2015repeatability}), we are more concerned by the lack of generality of enquiry, and consequently the generality of conclusions. We believe that, instead of examining individual datapoints concerning a particular algorithm implementation and a problem set of an author's choosing, we should be investigating deeper scientific questions, such as:

\begin{itemize}  
    
    \item[Q1] Why do our methods work? In particular, how can we assign credit to individual components and eliminate those that do not matter?
    
    \item[Q2] How can we \emph{analytically} arrive at a solution method given a problem description? For example, consider problem reduction: is it \emph{practically useful} to map an optimisation problem of a given type into another? Is there a ubiquitous ``SAT-like'' problem to which many optimisation problems can usefully be reduced to and solved in practice? 
    
    \item[Q3] Is it possible to more fully automate the exploration of the space of metaheuristics, allowing researchers to focus on improving that automation?
    
\end{itemize}

To these ends, this paper describes the vision and progress of `Metaheuristics in the Large' (MitL). MitL is a research community initiative (first introduced in Swan et al \cite{micstandards2015}) that seeks to address the lack of re-use at both conceptual and implementation level. MitL is both a synthesis and extension of existing ideas dispersed throughout the literature, and simultaneously a project producing new software tools and exemplars to show how these problems can be overcome. We draw on many contributions previously made in this direction: hyper-heuristics, constraint programming and the fundamental principles of substitutability of software components. We rely heavily on functional programming constructs to express metaheuristic components in a truly reusable way. In constructing this synthesis, we have exposed gaps in the literature that we are now closing with new contributions: in particular the MitL initiative has introduced a) the ``Automated Open Closed Principle'' \cite{AOCPECJ}, which shows how to express algorithm frameworks as `closed' design spaces which can nonetheless be configured in an open-ended manner via combinatorial assembly, and b) the removal of the domain barrier from hyper-heuristics \cite{recharacterizationhh}, essential in raising the level of genericity with respect to problem domains. These address Q3, making easier automation of the space of metaheuristics, and hence progress towards answers to questions Q1 and Q2.


\added{This paper provides a survey, progress report and roadmap of our attempts to reduce the fragmentation of metaheuristics research, improve reproducibility, and accelerate progress through infrastructure improvement. We  have made several concrete steps forward and can see the road ahead, but there are many problems left to be solved.}

\section{Contemporary Research Practice in Metaheuristics}
\label{sec:systemicissues}
One may characterize \emph{researchers} as being broadly concerned with the scientific practice of obtaining concise explanations of empirical observations \cite{popperconjectures}. In constrast, for \emph{practitioners} (e.g.\ in industry), the goal is to maximally exploit the information obtained via research, with minimal expert knowledge. We now present in more detail some of the challenges facing metaheuristic research, drawing on previous discussions in the literature. 
These observations motivate the MitL approach described in Section \ref{sec:approach}.

\subsection{Replication and Reuse}
Scientific progress in any discipline requires ready determination of the nature and merits of previous contributions, and the ability to build on the work of others to make further progress. Algorithm descriptions in many papers on metaheuristics are far from precise enough to allow independent re-implementation, and public access to the associated source code is rarely mandated by editors or programme committees. As a consequence, replication studies are very uncommon: a recent such paper, the only replication study in metaheuristics of which the authors are aware, obtained results which were an order of magnitude worse than those originally claimed \added{\cite{doi:10.1111/itor.12443}}\footnote{\removed{\cite{doi:10.1111/itor.12443}}It should be emphasized that this problem is not restricted to metaheuristics: a recent study~\cite{collberg2015repeatability} showed that in the computer science papers considered, 34.65\% of them were not repeatable and the authors could not conclusively determine repeatability in another 20.87\% of cases.}.

As noted above, metaheuristics, and in particular Evolutionary Computation research,
has developed something of a `throw away' culture
\cite{Pamp1512:Generic, collberg2015repeatability}, in which a large
percentage of researchers neither build upon the research
\emph{implementations} of their peers nor create such re-usable
software artifacts themselves. This inability to consolidate is in
contrast to other research areas that have successfully embraced
re-use: for example the SBML standard \footnote{\url{http://sbml.org}}
in systems biology, which allows the researcher to easily create test
and deployment pipelines \cite{Konig2020.01.04.894873}; or the Taverna
framework\footnote{\url{https://taverna.incubator.apache.org}} used
for workflow construction in a variety of other scientific
disciplines. One might wonder why metaheuristics, which enjoy a small
and ubiquitous set of abstract components such as acceptance or
perturbation, has seen relatively little progress in large-scale
re-use. It is possible that the very simplicity of metaheuristics at
the component level is in part responsible for this culture of
Babel-like proliferation. Metaheuristic researchers or practitioners
often choose the creation of \emph{ad hoc} solutions to using
pre-existing resources, perhaps because implementing baseline versions
of (say) Simulated Annealing or Genetic Algorithms from scratch is
relatively simple --- provided they are not intended for reuse by
others.

The last few decades have seen the development of many popular metaheuristic libraries, implemented in a variety of programming languages, some of which feature components that are (in principle) reusable at the framework level. Taking Evolutionary Algorithms as an example, Parejo et al.\ \cite{Parejo2012} provide an overview of commonly used libraries along with their performances on benchmarks, including
HeuristicLab~\footnote{\url{https://dev.heuristiclab.com/trac.fcgi/wiki/}} \cite{Wagner2005}, 
ECJ~\footnote{\url{https://cs.gmu.edu/~eclab/projects/ecj/}} \cite{Luke:2017},
FOM~\footnote{\url{http://www.isa.us.es/fom/}} \cite{Parejo2003},
Opt4J~\footnote{\url{http://opt4j.sourceforge.net/}}~\cite{opt4jpaper},
jMetal~\footnote{\url{http://jmetal.sourceforge.net/}} \cite{Durillo2011760} and
JAMES~\footnote{\url{http://www.jamesframework.org/}} \cite{DeBeukelaer:2017}. The majority of these libraries support component interoperability within their frameworks. However, a component implemented in a specific framework  cannot readily be reused within, or hybridized with, another framework. Recognizing this problem, an early attempt~\cite{guervos2003specifying} sought to achieve interoperability through the use of a common description language based in XML, albeit restricted in focus to evolutionary algorithms. PISA \cite{bltz2003a} was another early attempt to achieve interoperability \emph{across} frameworks\footnote{\url{http://www.tik.ee.ethz.ch/pisa/}}. In PISA, the problem domain component is separated from the metaheuristic component, and implementations of those components are reusable and interoperable, communicating via a file-based textual description. 

In practice the extensibility of these frameworks is limited (though uniquely, to our knowledge, a progression towards mechanisms that enable wider re-use can be seen in CIlib \cite{1501612,DBLP:conf/ijcnn/PamparaEC08,DBLP:conf/ijcnn/CloeteEP08, Pamp1512:Generic}), and, crucially, implementation often requires the modification of internal source code, presenting a barrier to distribution, reuse, and understanding for other practitioners. We are left with a fragmented set of implementations that are incapable of representing an extensible design space for metaheuristics, without requiring modification to the frameworks themselves. 

\subsection{Transparency}
Metaphorically-inspired approaches have recently suffered strong criticism for their lack of rigor. Where the use of metaphor obscures specific solution-domain mechanisms \cite{Sor13} the novelty of the metaphorical contribution becomes difficult to determine. At worst, this can lead to the re-invention or renaming of mechanisms that are already well-understood. For example, it has been argued that the popular `Harmony search' metaheuristic can be formulated as a simple variant of the foundational `Evolution Strategies' approach \cite{Weyland:2010}, and it has recently been claimed \cite{DBLP:journals/swarm/Camacho-Villalon19} that the `Intelligent Water Drops' algorithm is similarly not novel. Such `explanation by metaphor' unnecessarily obfuscates the field and makes it appear impenetrable to outsiders. 

This problem is at least partly cultural: the `reward' of publications and citations in metaheuristics is often more readily achieved by producing a method that `beats the competition' than one that makes the additional effort to be transparent about the contribution of its mechanisms \cite{Adriaensen:2014:FIS:2576768.2598285}. While improving on the state of the art should always be a key driver for a research community, the relentless pursuit of (apparent) novelty and the `up-the-wall game' \cite{Sor13} is counter-productive. For as long as researchers continue to labor in relative isolation, the risks of overfitting and misidentifying novelty remain present. In contrast, we propose in the following sections a more `bottom up' approach. With such an approach, new solution methods can be grounded in the principled decomposition of existing ones \cite{Lopez-Ibanez:2014:TDS:2598394.2609846}, thereby allowing ready identification of potential novelty. 

\subsection{Knowledge Discovery}
Hooker~\cite{hooker:95} and S\"orensen~\cite{Sor13} argue that there needs
to be more scientific analysis of how metaheuristics solve
problems. If a metaheuristic claims to work in some way --- say, for
example, it is claimed that a particular operator works by moving the
search out of local minima --- then experiments should be performed
that test this, or (even better) a theoretical justification
provided. This is a particular problem for complex metaheuristics,
where a number of innovations are often introduced in
tandem. Compounding this issue, the existence of `No Free Lunch'
theorems for optimization \cite{Wolpert:1997:NFL:2221336.2221408}
implies that metaheuristics often require domain-specific heuristics
for success \cite{recharacterizationhh}, and current practices mean
that any expert knowledge on which mechanisms work well on a given problem must typically be reverse engineered from publications on a per-case basis. To move to a more generalized level of enquiry, it is necessary to combine, exchange, and reason about metaheuristics and their component parts (such as acceptance, selection or perturbation) on a far larger scale than has been possible to date. Significantly, we believe that this requires a shift in community culture from `individual competition' to `collective knowledge discovery' and the development of a large pool of shared experimental data from which to draw general conclusions. 

One approach to deriving such general conclusions from a large pool of data is the application of data mining and machine learning (ML) techniques successfully used by other communities (e.g. meta-learning \cite{pappa:2014:contrasting}). Individual publications have applied ML methods to selecting or constructing heuristics, and provide some evidence for their efficacy: \added{Xu et al. \cite{xu2008satzilla} proposed a portfolio solver that won several SAT competitions by automatically selecting between various state-of-the-art SAT solvers 
based on a learned model of their relative performance conditioned on problem properties;} Thabtah and Cowling \cite{Thabtah:08} show associative classification can indicate which heuristic to use in each iteration of a personnel scheduling problem; Miranda et al.\ \cite{MIRANDA2017340} used fitness landscape information to decide whether to build or select a new particle swarm optimization algorithm; Nallaperuma et al. \cite{Nallaperuma2014,Nallaperuma2015} generated predictive models of the best parameters from ant colony optimization methods based on  features of previously evolved instances; Malan and Engelbrecht \cite{Malan2014} used landscape characteristics to predict the success of a collection of PSO algorithms on unseen continuous optimization problems;  Consoli et al.\ \cite{Consoli2014} used online learning and features extracted from the fitness landscape of the problem to choose the most appropriate genetic operator and Asta et al.\ \cite{Asta:2015, Asta:2016} integrated knowledge discovery directly into the search algorithm. 

More generally, Smith-Miles et al.\ \cite{SmithMiles201412} proposed a methodology where instances of a problem are represented by a set of features in an instance space, and machine learning algorithms used to classify the regions of the space where algorithms are expected to perform well or poorly, given many insights on algorithms strengths and weaknesses. Such feature-based approaches provide a baseline for generating and mining knowledge of relevance to metaheuristic research and practice.

The re-use of such knowledge will first require a knowledge-base and the associated effort to constantly update it. Initial efforts towards a schema for such a database was presented by Scheibenpflug et al
\cite{Scheibenpflug:2012:OKB:2330784.2330806}: their Optimization Knowledge Database (OKD), contained data about algorithms, the problems they were used to solve and their parameters. The authors emphasize that populating the database is time-consuming and requires the effort of the whole community. Other works in the literature have created specific instances of such datasets \cite{MIRANDA2017340,SmithMiles201412}. Such a community effort to effectively create and populate a knowledge base is paramount for the success of metaheuristics mining. Given a sufficiently rich representation for components, such analysis could be carried out in a semi-automated way. 

\subsection{Automated Design}
Contemporary scientific and engineering disciplines rely heavily on standardization and automated tools. The design of these tools and their underlying algorithms tends to be an \emph{ad hoc} process, often regarded as an art rather than a science \cite{hunt2001art}. As a consequence, the design of an algorithm is time-consuming and costly. Furthermore, the process itself is rarely documented, making it untraceable, i.e.\ it is often unclear what motivated certain design decisions (e.g. expert knowledge, experimentation, intuition) and which alternatives were considered. Not only do we lose potentially interesting information and insights which can be used to design algorithms in the future, it also makes the process susceptible to accidental human bias.

The automated design of algorithms has significant potential to address these issues. Unsurprisingly, attempts to (partially) automate algorithm design, in one form or the other, are ubiquitous and can be traced back to the origins of computational intelligence (e.g.\ program synthesis \cite{manna1980deductive}, genetic programming \cite{koza1992genetic}, swarm algorithms \cite{Khichane2008a}, algorithm selection \cite{rice1976}, algorithm configuration \cite{Birattari2002irace}). However, the application of these techniques has thus far been largely a privilege of experts, restricted to isolated case studies, and is far from a standard practice in algorithmics.

\removed{One could argue that the computer science community is often somewhat reluctant to turn its own methods towards the development of computer science research itself, and t}
\added{T}he metaheuristics community is no exception. \removed{While metaheuristics researchers are enthusiastic about applying search and optimization strategies to a vast wealth of problem areas, it is only in recent years with the emergence of generative hyper-heuristics \cite{Burke2013} that these methods have been applied to metaheuristics research itself.}
\added{While metaheuristics are most commonly designed manually, the idea of automating this process is hardly new, and has been actively pursued for almost two decades in the hyper-heuristics~\cite{Burke2013} and the algorithm configuration \cite{stutzle2019automated} communities.}
An important aspect of our vision (see Section \ref{sec:ourvision}) is to facilitate the integration and further development of these design automation techniques.

\subsection{Scalability}
Historically, computing systems have tended to get faster at an exponential rate. Software performance automatically scaled along, without requiring any additional efforts from the developer. The situation is no longer so simple: contemporary systems, rather than getting faster and faster, are able to do more and more work in parallel \cite{sutter2005free}. To take advantage of increasing parallel processing capabilities, computations must be subdivided into a set of interdependent tasks to be executed efficiently in parallel across multiple cores and/or across networked machines. In computer science in general, much human effort has been invested in algorithm-specific parallelization strategies. 

Scalability is also an issue when solving ever larger problem
instances: despite the increase in computing power, it is hard to
solve large instances of many practical optimization problems. This
will of course always be the case --- contrary to other computational
domains, the field of combinatorial optimization will never have
``enough'' computing power. Fortunately, many popular metaheuristics
are `embarrassingly parallel': for example, determining the fitness of
each population member in evolutionary approaches can be readily
parallelized; strictly from the performance point of view and
depending on latency and throughput, this simplistic approach might
not be the most efficient; however, the fact that it can be done at
all shows that there are parallel approaches which are functionally
equivalent to sequential ones and that have a straightforward implementation.

Currently many metaheuristic methods rely on parallelism at a specific
level of abstraction, typically by parallelizing either fitness
evaluation or part of a population of solutions using an island
model. Both these approaches are limiting in the assumptions they make
about the complexity of metaheuristics: to achieve sophistication
beyond previous applications may require much more complex and
involved search operators, for example, and we may wish to parallelize
a metaheuristic not just at these fixed and algorithm-specific levels,
but to the greatest extent possible, i.e. at the level of individual
components; in this sense, it might also be convenient to simply use
underlying concurrent or parallel models, such as communicating sequential processes \cite{DBLP:conf/gecco/GuervosVG20}.

\section{The `Metaheuristics in the Large' Approach}
\label{sec:approach}
The challenges of the previous section motivate the creation of infrastructure support for community-wide sharing of problems, metaheuristic frameworks and heuristic components. In general, there are many good tools and libraries already in existence. We do not propose to reinvent or replace them; we are not proposing ``just another library'' but rather a different way of structuring and implementing metaheuristics research.
~%
\newline
~
\newline
Our proposed approach has three conceptual underpinnings:

\begin{itemize}
\item \textbf{Extensible and re-usable framework templates} \\
To support open-ended innovation and provide true reusability, these templates (or any other suitable problem description language) must be configurable via a palette of components that is extensible \emph{without requiring the modification of existing code}. To support such extensibility and the automated configuration of these templates, we require a stronger notion of interoperability than existing software: there must be infrastructure support for \emph{state-threading}, i.e.\ passing framework or component-specific state (such as the temperature in simulated annealing) via a dedicated mechanism. This strong notion of extensibility is described in Section~\ref{automatedassembly}. Our approach facilitates reuse through this extensibility, and transparency and automated design by explicitly encapsulating component behavior (rather than relying on metaphorical description) and allowing machine-inspection of behaviors.

\item \textbf{White box problem descriptions}\\ 
Having embraced the necessity of state threading, it follows that we can thread more than merely empirically-obtained data relating to the search trajectory. In particular, threaded state can include \emph{analytic} information, such as declarative/whitebox problem descriptions. This analytic information can be used to guide algorithm selection or construction in a more informed manner than has traditionally been embraced by the hyper-heuristics community \cite{recharacterizationhh}. We discuss this further in Section \ref{whitebox}.

\item \textbf{Remotely accessible frameworks, components and problems}\\
By building upon the two concepts above, it is possible to configure pre-existing, remotely-hosted, algorithm frameworks with some (potentially newly-devised) collection of heuristic components. The practical obstacle to further progress is then the relatively procedural one of community agreement on definitions for component interfaces and communication protocols. For inspiration, we look to work on `Service Oriented Architecture', which we discuss in Section \ref{SOA}. This enables widespread reuse, replicability, and shared knowledge discovery.
\end{itemize}

\subsection{Re-usable Framework Templates\label{automatedassembly}}

The main obstacle to the open-ended extension and automated composition of existing implementations of metaheuristic components is that they suffer from an intrinsic lack of modularity. In this section, we illustrate why this is an issue for research `in the large' and describe the proposed solution. In part, this is due to the lack of adoption of best practice from software engineering \cite{DBLP:conf/gecco/GuervosVG20}. In the case of metaheuristics, there is an added complexity due to  state dependencies between different algorithms components, which we now describe.

Framework configuration can be defined in general terms by expressing frameworks as higher-order functions that take components as parameters. For example, a possible \added{function} signature\added{\footnote{\added{The \emph{signature} of a function is the formal description of its parameter and return types.}}} for acceptance \replaced{that is type parameterized by}{for some} generic candidate solution \lstinline$Sol$ is:
\[\replaced{accept_{Sol}}{accept}: \mathrm{incumbent}: Sol \times \mathrm{incoming}: Sol \rightarrow \operatorname{Boolean}\]

Listing \ref{simplels} gives a simple local search framework  that allows for three design decisions, viz.\ the choice of perturbation, acceptance and termination conditions. \added{In order to support alternative designs, local search is a \emph{higher-order function}: it takes as arguments separate \emph{functions} for \emph{perturb}, \emph{accept} and \emph{finished} and returns a candidate solution of type Sol. As described above, \lstinline$accept$ takes as argument a pair of candidate solutions (i.e.\ the incumbent and incoming solutions) and returns the prefered one, denoted in the following listing as: 
\lstinline$accept: (Sol,Sol) => Sol$. \lstinline$perturb$ and \lstinline$finished$ are defined the correspondingly obvious manner.}

\begin{lstlisting}[float=htb,mathescape=true,caption=Local Search framework parameterized by design choices, label=simplels]
def localSearch(
      incumbent: Sol, 
      perturb: Sol => Sol, 
      accept:  (Sol,Sol)  => Sol, 
      finished: Sol  => Boolean
    ): Sol {

    while( not finished( incumbent ) )
        incumbent = accept( incumbent, perturb( incumbent ) )

    return incumbent;
}
\end{lstlisting}

Each specific triple of components (\emph{perturb}, \emph{accept},  \emph{isFinished}) used to configure the framework corresponds to a specific local search algorithm. This allows us to concisely specify a combinatorial design space of alternative component configurations, and also makes design space commonalities explicit. In order for a framework to permit the substitution of different choices for each component, components must ultimately conform to some well-defined interface, e.g.\ the higher-order function arguments to the framework have some signature that is fixed \emph{a priori}. In our example, for candidate solution type \emph{Sol} (e.g.\ a list of cities in the Traveling Salesperson Problem) perturbation is assumed to have type $\text{\emph{Sol}} \rightarrow \text{\emph{Sol}}$. 

However, such fixed signatures are problematic if we wish such frameworks to be `closed to modification', i.e.\ be able to accommodate unanticipated component dependencies without requiring changes to framework code. The need for such modification is clearly incompatible with the MitL goal of frameworks that can be both shared across the research community and be configured (perhaps automatically) with new components. As a concrete example: suppose we now wish to incorporate a further heuristic that requires information about the search trajectory, e.g.\ a tabu list of solutions \citep{Glover:1997:TS:549765}  that promotes search diversification. We are therefore required to change the implementation of local search to keep track of the trajectory. Listing \ref{simplelshistory}, gives a revised version in which the history list of previous incumbent solutions is denoted by $[\text{\emph{Sol}}]$.

\begin{lstlisting}[float=htb,mathescape=true,caption=Explicit incorporation of solution history,label=simplelshistory]
def localSearch( 
    current: Sol, 
    history: [Sol], 
    perturb: (Sol,[Sol]) => Sol, 
    accept: (Sol,Sol,[Sol]) => (Sol, [Sol]),
    finished: (Sol,[Sol]) => Boolean 
    ): (Sol,[Sol]) {

    while( not finished( current, history ) )
        (current,history) = accept(current,perturb(current,history),history);

    return (current,history);
}
\end{lstlisting}

The modified implementation now supports solution-based tabu mechanisms, but the issue of course persists if we wish to incorporate components which require new state dependencies, for example Metropolis-Hastings acceptance, which requires some measure of `temperature' \citep{DBLP:journals/science/KirkpatrickGV83} to be statefully maintained. In the general case, we clearly cannot anticipate in advance what information will be required by some component yet to be devised. These are examples of \textit{environmental state}, which provides the context for decisions made by the search process. For extensibility, it is therefore necessary for support for environmental state to be open-ended, i.e.\ for frameworks to be configurable with components that access aspects of environmental state that are not known at the time of framework implementation. Principled handling of environment state is key to metaheuristic modularization, and is therefore essential for both component interoperability and scalability in automated construction of metaheuristics. \added{The technical specifics of MitL support for this approach are described in an associated publication \cite{AOCPECJ} and summarized in \ref{sec:automatedopenclosedprinciple}. Software exemplars of the proposed infrastructure support are publicly available, as described in \ref{sec:mitlsoftwarelibraries}.}

\msgtoreviewer{The remainder of this section was moved to \ref{sec:automatedopenclosedprinciple}. Changes are highlighted there.}

\subsection{White box Problem Representations\label{whitebox}}
It is well-known that the exploitation of problem information is key in rendering optimization problems tractable \cite{Wolpert:1997:NFL:2221336.2221408}. As such, a principal challenge lies in devising frameworks and solvers that support injection of problem information to drive the search process, without incurring loss of genericity (i.e.\ they can be applied to many different problems).

By including \emph{white box problem descriptions} as part of the environmental state, it is possible to define frameworks in terms of rich domain information, allowing the aforementioned challenge to be tackled using an open-ended combination of human ingenuity and automation.

\subsubsection*{What Kind of Information Can Be Exploited?}
In principle, \emph{any} machine-readable knowledge could be exploited to bias the search, to synthesize feasible operators, etc. As discussed in the introduction, this knowledge can be split into two categories:
\begin{description}
\item[Analytic] knowledge about intrinsic features of the problem. 
\item[Empirical] knowledge gained through experience, i.e.\ experimentation. 
\end{description}

\noindent
Many examples of successfully exploiting a combination of analytic and empirical knowledge can be found in the literature:

\begin{itemize}

    \item Reactive tabu search \cite{battiti1994reactive} exploits knowledge about the presence of specific substructures in candidate solutions to diversify the search, and uses trajectory information to adapt the tabu tenure parameter dynamically.

    \item Variable neighborhood descent \cite{hansen2001variable} exploits knowledge about the relative sizes of multiple domain-specific neighbor relations to (local) search them more efficiently, and uses empirical information (a candidate solution being locally optimal/improving) to switch between neighbor relations.
    
    \item Matheuristics (e.g.\ \cite{Addis2013,Nikzad2021} typically combine analytical approaches such as ILP to solve sub-problems, with higher-level searches using empirical feedback on solution quality to build the results into solutions for a larger-scale problem.
    
    \item Solution spaces can be decomposed using techniques with their origins in mathematical programming, to increase efficiency at the metaheuristic level \cite{Raidl2015}.
    
    \item Constraint relaxation (e.g.\ \cite{Fuellerer2010}) typically uses analytical knowledge of the acceptable constraint bounds to allow a metaheuristic to search across infeasible regions of the space using empirical feedback on solution quality to determine when the relaxation should be reduced. Similarly, different heuristics can be targeted at different constraints, driven by analytical knowledge of the constraints themselves \cite{GOH201717}.
    
    \item Portfolio solvers (e.g.\ SATzilla \cite{xu2008satzilla}), select between multiple solvers based on analytic features of the problem instance to be solved. The mapping from features to solvers is generated empirically, using machine learning. 

\end{itemize}

\subsubsection*{Historical Development Towards White Box Approaches}
While the importance of exploiting problem structure is widely recognized, arguably there is a historical aversion to do so at the hyper-heuristic level, leaving this task up to the domain-specific instantiations or low-level heuristics \cite{drake:2019:survey}. Maintaining generality is often cited as the motivation for this information hiding practice. For example, Chakhlevitch and Cowling \cite{Chakhlevitch2008} argue for the importance of limiting problem domain information in achieving cross-domain generality in selection hyper-heuristics. They argue that a framework can be applied to any problem that shares the ``lowest common denominator'' characteristics. While sufficient for generality, information hiding is not necessary. It is easy to see that a framework can exploit arbitrary information without loss of generality, as long as it is also capable of solving the problem without it. For instance, a general optimizer could use gradient information when available (e.g.~when training neural networks) and default to a derivative-free approach otherwise.

The progression of hyper-heuristic research demonstrates an increased acknowledgment that use of white box problem descriptions is both possible and desirable. Following the pattern set by initial work \cite{cowling:2001:HH}, most of the selection hyper-heuristics studies maintain a black-box interface between the hyper-heuristic and problem domain known as the {\em domain barrier}. The original rationale for the domain barrier, which disallows a hyper-heuristic from retrieving any problem-specific information, was thought to be necessary for cross-domain generality. However, it has been recognized that the domain barrier might be more a problem than a feature: Ross \cite{ross:2014:survey} argued that an explicit domain barrier that enforces a strict separation between the hyper-heuristic and the problem-specific aspects makes hyper-heuristics undesirable for use in large real-world applications. Furthermore, Parkes et al \cite{parkes:2015:software} and Pappa et al \cite{pappa:2014:contrasting} suggested an increased exchange of information between the problem domain and the higher search level which could then be analyzed via data science techniques and machine learning. More advanced learning for heuristic selection has progressively been introduced \cite{burke:2007:graph,qu:2009:timetabling,soria:2017:methodology,kheiri:2016:iterated,ahmed:2019:routing,BENGIO2020,AGARWAL2006801}. 

Recent work throws further doubt on the necessity of the domain barrier. Swan et al \cite{recharacterizationhh} state that work in constraint-satisfaction provides abundant evidence that problems can be described in a domain-independent manner without loss of solver generality. The lack of necessity for the domain barrier was further evidenced by Kheiri \cite{kheiri:2019:irp}, who designed a hyper-heuristic utilizing extended domain information that nonetheless manages low-level heuristics in a domain-independent manner. Martin et al \cite{Martin2016} designed a hyper-heuristic controlling the parameter settings of randomised heuristics based on an agent-based cooperative search framework. An ontology is used to translate problem-specific elements into problem-independent abstract objects. 

\subsubsection*{White Box Descriptions and Automation}
In order to communicate the required information to the solver and its heuristics, it is necessary to move beyond the bespoke approaches above: it must be possible to communicate arbitrary domain knowledge to an algorithm framework template. In a black box setting, life is simple: solvers can choose from a closed set of interfaces and can solve any problem that implements it. In contrast, the white box setting is completely open-ended: solvers can require whatever information they deem fit. Clearly, this presents an array of novel challenges. Most notably:
\begin{enumerate}
\item Deducing (analytically) which solvers can be applied to which problems.
\item Overcoming limited applicability due to interface mismatches.
\end{enumerate}
This is clearly a rich  topic for further research. However, we believe that some foundational aspects can be identified:
\begin{enumerate}
\item The use of a declarative, machine-readable language to express the information problems provide, solvers require, and their relations.
\item Automated algorithm selection and problem (re)formulation, as facilitated by white box descriptions \cite{recharacterizationhh}.
\end{enumerate}

The practical choice for such a language is rightfully an open-ended research question, but the MitL proposal \cite{recharacterizationhh} is that prior art in Constraint Programming provides a suitably generic baseline. Existing standards within Constraint Programming, such as XCSP3 \cite{DBLP:journals/corr/BoussemartLP16} support whitebox descriptions of constraints that capture a very wide range of common problems. As research advances, the set of supported constraints can further expand, without limiting applicability. Ultimately, interface boundaries will fade, causing a paradigm shift, where human researchers specify components using problem information and computers assemble them into a single framework, automatically selecting the component believed to work best, based on all available analytic and empirical information.

\subsection{Service Oriented Architecture \label{SOA}}
A large-scale solution to lack of re-use lies in ``Service Oriented Science'', which applies the increasingly-widely adopted practice of service-oriented architectures \cite{HACKNEY20061161} to scientific computing. The concept is defined as ``the pursuit of scientific research using distributed and interoperable services, the accessibility of these interfaces being the key to success'' \cite{Foster2005Science}. By such means, researchers can discover and access services without developing specific programmatic clients for each data source, or program. Such an approach clearly has the potential to increase scientific productivity via public and distributed services, and also to increase data analysis automation. There are many examples that attempt to boost this paradigm, such as the Open Science Grid \cite{Altunay2011OpenScience} and GLOBUS \cite{Foster2005Globus}. These projects include scientific communities and globally distributed infrastructures that support scientific and integrated applications of different domains.

As we have argued above, it is highly advantageous for metaheuristic researchers and practitioners to converge on a standard machine-readable language for problem description, experimental configuration and results. Service Oriented Architectures (SOAs) offer several ways to build a research workflow from these elements. SOA is a computational paradigm in which agents interact using loosely coupled, coarse-grained, and autonomous components called {\em services} \cite{rotem2012soa}. A {\em service} is a distributed entity, such as a node, program or function, used to obtain a result, increasing the integration of  systems that are heterogeneous in respect of operating systems, protocols or languages.

The SOA perspective promotes the creation of services that are discoverable and dynamically-bound, self-contained/modular, loosely-coupled, location-transparent and composable \cite{Valipour09surveysoa}. As such, SOA is clearly therefore a good fit for the process of ``devolved community research'' which we advocate here. Lately, SOA has seen a trend towards ``microservice architectures'': distributed, cloud-based and cloud-native, these architectures follow the principle of separation of concern to create applications that are easily scalable and deployable, with a stable response and maximum availability. Several frameworks, such as the one proposed by Khalloof et al.\ \cite{khalloof2018generic}, exploit the capabilities of microservices to create scalable systems that can be used at different levels (from the desktop to the web) for optimization. However, at the time of writing there are no generally accepted standards for microservice discovery, and although they offer some advantages in term of composability and scalability, they lack the service representation feature that would make it amenable to use within a large-scale metaheuristics framework.

\subsubsection*{SOA for Metaheuristics}
Previous work on SOA for metaheuristics has mainly been concerned with the application of a specific metaheuristic, such as Genetic Algorithms, to optimize a service selection or composition based on the QoS (Quality of Service) of their execution \cite{RosenbergMLMBD10}.

Different SOA technologies, such as web services, have been proposed for solving optimization problems via grid computing \cite{grid1,grid2,grid3}, where services are defined using WSDL (Web Services Description Language) interfaces and other transmission mechanisms (such as Remote Procedure Call \cite{grid8} or Globus Toolkit \cite{grid10}). ROS (Remote Optimization Service) \cite{GarciaNieto07ROS} was one of the first attempts to allow remote execution of metaheuristics, with inputs and outputs described via XML specification. Other metaheuristic frameworks such as HeuristicLab include plug-ins to allow parallelism and interoperability using web services \cite{HEURISTICLAB}. GridUFO is a service oriented framework \cite{gridUFO}, but it only allows the modification of the objective function and the addition of whole algorithms, without combining existing services. 

 Garc\'ia-S\'anchez et al \cite{GarciaSanchez13SOEA} have previously proposed a SOA for Evolutionary Algorithms. Several suggestions on different concerns about the design and development of the elements of an EA using SOA were presented, such as the operator behavior, dynamism or solution representation. A specific SOA technology, OSGi, was used as an example of implementation. More recently, MOSES \cite{Parejo15} was proposed as the design of a global architecture based on service contracts, allowing the automation of the experimentation process in a metaheuristic optimization context. This architecture is based on different tools, such as specific automatic experimental description languages and statistical services, forming a contract-based chain of software components for experimental execution. 
These global architectures have already been proposed for other fields, such as distributed simulation \cite{TAYLOR20191}. 

Most of these approaches are more or less direct mappings from the original implementation to a SOA framework; however, one key way in which the proposed approach facilitates knowledge discovery is the ability to add arbitrary instrumentation to components via the generic environment representation. In particular, this allows for data mining on metaheuristic traces. In addition, by employing our generic notion of state (which denotes one or more solutions, together with any environmental information required to represent the current algorithmic state of the search), the same framework can instantiate metaheuristics operating at different scales. For example, a composite recombination operator can choose from different types of recombination strategies, putting meta and hyper-heuristics under the same framework (e.g.\ in the manner of \cite{Woodward:2014:CDP:2598394.2609848}). Having these different types of algorithms under a common framework greatly facilitates their extension and comparison.

More generally, we envision the emergence of a distributed, community driven suite of tools, providing an expanded repository of interoperable frameworks and components, bringing together researchers and practitioners across domains, unifying the field and closing the gap between scientific research and empirical practice.

\section{\removed{Progress towards MitL}\label{sec:progress}}
\msgtoreviewer{This section was moved to  (\ref{sec:mitlsoftwarelibraries}). Except for the last paragraph which was moved to next section.}

\removed{As we have intimated, a purely technical solution to the issues of metaheuristics research is insufficient: community-level engagement is also required. While
we have proposed a means by which extensible algorithm templates can integrate with other frameworks, the ultimate arbiter for success is the enthusiasm
of the wider research community to embrace such initiatives. In the following
section, we describe some of the prospective benefits of doing so.}

\section{Use-Cases for MitL}

\label{sec:ourvision}

\added{As we have intimated, a purely technical solution to the issues of metaheuristics research is insufficient: community-level engagement is also required. While we have proposed a means by which extensible algorithm templates can integrate with other frameworks, the ultimate arbiter for success is the enthusiasm of the wider research community to embrace such initiatives. In this section, we describe some of the prospective benefits of doing so.}

A recent paper by Kendall et al.\ \cite{Kendall2016} has, importantly, emphasized the need for good laboratory practice in optimization research. The set of practices that they advocate include making datasets available in a standardized format, reporting the results from the individual components of a hybrid approach, describing in a reproducible way the evaluation function and the metaheuristic used, clearly presenting computational times, the use of appropriate statistical tests, etc. We disagree with none of this. However, a driving philosophy for MitL is that it is necessary to go beyond the mere \emph{advocacy} of good practice, to making it easy --- indeed, almost inevitable --- that good practice can happen. The MitL proposal is that it is possible to embed foundational support for good practice directly into the software that is used by metaheuristics researchers, consequently making the fruits of that research available to other practitioners as via their default workflow. Below, we describe some specific use cases in metaheuristic research which are facilitated by the proposed approach.

\subsection{Comparison between Metaheuristics}

Many papers in the metaheuristics literature compare the performance 
of a new metaheuristic against a small sample of other
metaheuristics. Sometimes, it is made clear that the comparators
chosen have been specifically selected because they represent the
state-of-the-art for that particular problem area: however, in many
cases this is not made clear. Furthermore, many papers simply compare
the new metaheuristic against other metaheuristics of the same broad
type, for example, comparing a new variant on Particle Swarm
optimization (PSO) against other PSO variants.

Providing some evidence for the effectiveness of a new method is
clearly important. However, as the number of metaheuristics continues
to expand, comparing against a few other metaheuristics seems weak;
even where an assertion is made that the comparators represent the
state-of-the-art, this is usually presented as an assertion to be
taken on trust, and the method used to choose the comparators is
unstated. In particular, there is no guarantee that the chosen problem instances actually exhibit different landscape characteristics. We therefore propose that
creators of new metaheuristics should test their metaheuristics
against \emph{all} other appropriate metaheuristics.

The overall aim is therefore to ensure that comparisons are
both thorough and fair. This is clearly only realistic at a 
large scale if the process of comparisons is automated. 
Some initial progress in this direction was made several years ago: Taillard et al.\ \cite{Taillard2005} suggest that iterative metaheuristics should be compared not only for a single computational effort (e.g.\ giving the best solution found after a fixed number of iterations), but also continuously at each iteration\footnote{\url{http://mistic.heig-vd.ch/taillard/qualopt/}}. Some recent work has also started to address the issue of fair 
comparison of algorithms by providing statistical testing 
frameworks which ensure that the preconditions for the various 
tests applied are actually met \cite{Neumann:2014:EET:2598394.2609850}. 
This is particularly important for metaheuristics, since 
common assumptions (e.g.\ of normality) are not in general true. 
There is also a need to ground reported results in terms of 
``effect magnitude'' \cite{Neumann:2014:EET:2598394.2609850}: 
for example, an improvement of 0.1\% on the state-of-the-art may have more practical relevance for the Traveling Salesman Problem than for Bin-packing. In addition to statistical considerations, the specifics of the termination condition are obviously also a vital aspect of fair comparisons. We claim that the transparency afforded by the proposed approach is vital in ensuring that comparisons are commensurate.\footnote{For a community effort that promotes best practices in benchmarking, we refer the curious reader to~\cite{Bartz2020benchmarking}. It discusses eight topics: clearly stated goals, well-specified problems, suitable algorithms, adequate performance measures, thoughtful analysis, effective and efficient designs, comprehensible presentations, and guaranteed reproducibility. }

\subsection{Testing Against Problem Instances}

\noindent
Establishing the effectiveness of a metaheuristic requires two components: 
other metaheuristics against which to compare, and a range of problem instances
upon which to compare them. Again, many papers in the
metaheuristics literature are considerably less than comprehensive in the 
range of problems which are investigated. While certain benchmark test suites 
are available, the application of a specific new method to these
benchmarks is often done in a seemingly \emph{ad hoc} way, with a number of 
examples from the benchmark chosen without any justification \cite{hooker:95}.

As described in previous MitL work \cite{recharacterizationhh}, progress in this area depends on the ability to express a wide variety of problem types in a common format.  This further facilitates the creation of systems that could apply a new metaheuristic systematically over a wide range of problems. Researchers would not be limited by the amount of time it would 
take to set up experiments with a large number of problem instances; 
an automated script can work through a repository 
of problems, automatically applying the new metaheuristic to each appropriate  example. 

Of course, \added{benchmarking metaheuristics cannot be reduced to} simply counting the number of problems for which a
particular metaheuristic is ``better'' than another one \replaced{is meaningless, since it is well
known (via the No Free Lunch theorem) that there is no “universal” metaheuristic \cite{Wolpert:1997:NFL:2221336.2221408}}{\cite{Bartz2020benchmarking}}.
Rather, the aim is to gather a rich set of data about the 
performance of each metaheuristic on a wide variety of problem 
types and instances. \added{Such knowledge database, in combination with white box descriptions of problems (and introspectable methods), could then be mined to gain deeper insights into which methods work best when (and why). In particular, it allows more general metaheuristics to be constructed automatically from more specialized ones (e.g., using algorithm selection portfolios as in \cite{xu2008satzilla})}.


\subsection{Hybridization}

The hybridization of solution methods has been a successful approach for combining the complementary strengths of different optimization paradigms and to reduce their individual weaknesses with the aim of obtaining more effective algorithms (see e.g., \cite{CPAIOR-tenyears} for a review from the perspective of the CP-AI-OR community). Hybrid approaches can be classified according to many dimensions \cite{Puchinger2005}, e.g.\ whether the components involved in the hybridization come from different search paradigms (usually constructive methods or exact methods such as Constraint Programming or Integer Linear Programming) or whether they are homogeneous (e.g., local search or evolutionary methods). Indeed, presenting a specific 
hybridization of two or more metaheuristics is a common source of novelty in metaheuristic research. Unfortunately, the way in which these hybridizations are evaluated is often a very simplistic comparison in terms of accuracy or error measures, without any attempt to 
attribute specific behaviors in a run of the metaheuristic to particular components, or indeed to perform any elimination of ``accidental complexity'' \cite{Adriaensen:2014:FIS:2576768.2598285}.

Often, the way in which metaheuristics are combined is the strongest contribution of a method. For example, 'Fair Share ILS' \cite{Adriaensen:2014:FIS:2576768.2598285} performs well because of the synergistic interaction between its acceptance and perturbation heuristics. More generally, it is of relatively little use for perturbation to be operating strongly as a ``search intensifier'' if the acceptance criterion only permits large increases in solution quality. Such decisions are best informed via large-scale studies \removed{of} (as supported by combinatorial assembly over a range of problems and algorithm configurations) and component instrumentation (as supported via the proposed environmental state threading).

\subsection{Matching Metaheuristics to Problem Types}

Although there has long been interest in relating problem characteristics to solution strategies \cite{rice1976}, we claim that this is an area that has been particularly hindered by the lack of re-use. One problem with these studies, valuable as they are, is that they represent a single sample point in time. As new metaheuristics are
created, the value of that cross-cutting analysis becomes weaker, as
researchers present new, \emph{ad hoc} evidence for the value of a particular metaheuristic on a particular problem. Given the importance of such cross-cutting studies, we propose that it is key for the community to support a constantly-updated repository of
metaheuristics and experiments, and subsequently so that the most effective metaheuristic for a particular problem area can be identified and kept up-to-date. Importantly, this would need more than just the creation of such a
repository. Analysis tools would also be needed, which would mine the
ever-expanding repository to find features that best predict which
kind of metaheuristic is well-suited to a novel dataset. This would ideally involve 
the automated application of metaheuristics to problems, with the repository constantly being updated as new problems and (meta)heuristics are added.



\section{Conclusion}
\label{sec:conclusion}

Metaheuristics in the Large (MitL) is a community project that addresses some of the cultural and technical issues we believe are impediments to progress in metaheuristic research: 

\begin{itemize}

    \item Through the MitL component-based architecture and explicit state threading outlined in Section \ref{sec:approach}, heuristics can be described from a behavioral standpoint, moving away from an over-reliance on metaphor and the accidental re-invention of established heuristics.

    \item MitL eliminates the need to modify existing framework source code when implementing new heuristics. Whilst this is of great benefit to a practitioner, it is \emph{essential} for the open-ended combinatorial assembly of metaheuristics. Design automation can raise the abstraction level of research from manual labor such as parameter tuning and selecting and combining heuristics, towards answering more general scientific questions.
    
   \item Within the space of a few years, Deep Learning approaches have changed the perspective on what is possible in Machine Learning. By following MitL's approach to defining heuristics, it should be easier for practitioners to recursively define very large-scale metaheuristic architectures (e.g.\ exploiting parallelism) without undue concern for low-level implementation details, enabling exploitation of the large-scale parallelism of modern compute platforms.

\end{itemize}

The approaches described in Section \ref{sec:approach} combine to provide a basis for extensible Software as a Service implementations of metaheuristics provided via stateless web-services, supporting shared framework templates which allow combinatorial assembly and comparison of metaheuristics. The language and platform agnosticism of this approach in turn addresses issues of reproducibility and scalability.

Metaheuristics are one of the great contributions to practical computer science of the last few decades. However, without interoperable frameworks for analyzing, comparing and hybridizing them, advances in the \emph{science} of metaheuristics are few and far between. Once such frameworks are in place, we will be able to put metaheuristics on a much more experimentally rigorous footing, to advance the science of metaheuristics, and to build a communal resource that is of benefit to both practitioners and researchers in this important area of computational intelligence.

In this article, we have described the key functionality that supporting infrastructure  requires. What is now needed is community consensus on the relatively procedural aspects of interoperability protocols. Editors and reviewers can then insist on a thorough and systematic application of new metaheuristics to a wide range of problems, with the attendant rich 
analysis possibilities that are opened up. 

\subsubsection*{Acknowledgments}
Many people have generously given their time to the various activities of the MitL initiative. Particular gratitude is due to
Adam Barwell, John A.\ Clark, Patrick De Causmaecker, Kevin Hammond, Emma Hart, 
Zoltan A.\ Kocsis, Ben Kovitz, Krzysztof Krawiec, John McCall,
Nelishia Pillay, Kevin Sim, Jim Smith, Thomas St\"utzle, Eric Taillard and Stefan Wagner.

J.\ Swan acknowledges the support of UK EPSRC grant EP/J017515/1 and the EU H2020 SAFIRE Factories project. P.\ Garc\'ia-S\'anchez and J.\ J.\ Merelo acknowledges the support of TIN2017-85727-C4-2-P by the Spanish Ministry of Economy and Competitiveness. 
M.\ Wagner acknowledges the support of the Australian Research Council grants DE160100850 and DP200102364.

\vspace{0.3cm}
\noindent\msgtoreviewer{While the changes in the bibliography are unfortunately not highlighted, we did perform the requested changes.}

\small
\bibliographystyle{apalike}
\bibliography{towards_mitl}

\begin{thebibliography}{}

\bibitem[Addis et~al., 2013]{Addis2013}
Addis, B., Carello, G., and Ceselli, A. (2013).
\newblock Combining very large scale and ilp based neighborhoods for a
  two-level location problem.
\newblock {\em European Journal of Operational Research}, 231(3):535 -- 546.

\bibitem[Adriaensen et~al., 2014]{Adriaensen:2014:FIS:2576768.2598285}
Adriaensen, S., Brys, T., and Now{\'e}, A. (2014).
\newblock {Fair-share ILS: A Simple State-of-the-art Iterated Local Search
  Hyperheuristic}.
\newblock In {\em Proceedings of the 2014 Annual Conference on Genetic and
  Evolutionary Computation}, GECCO '14, pages 1303--1310, New York, NY, USA.
  ACM.

\bibitem[Agarwal et~al., 2006]{AGARWAL2006801}
Agarwal, A., Colak, S., and Eryarsoy, E. (2006).
\newblock Improvement heuristic for the flow-shop scheduling problem: An
  adaptive-learning approach.
\newblock {\em European Journal of Operational Research}, 169(3):801 -- 815.

\bibitem[Ahmed et~al., 2019]{ahmed:2019:routing}
Ahmed, L., Mumford, C., and Kheiri, A. (2019).
\newblock Solving urban transit route design problem using selection
  hyper-heuristics.
\newblock {\em European Journal of Operational Research}, 274(2):545--559.

\bibitem[Altunay et~al., 2011]{Altunay2011OpenScience}
Altunay, M., Avery, P., Blackburn, K., Bockelman, B., Ernst, M., Fraser, D.,
  Quick, R., Gardner, R., Goasguen, S., Levshina, T., Livny, M., McGee, J.,
  Olson, D., Pordes, R., Potekhin, M., Rana, A., Roy, A., Sehgal, C., Sfiligoi,
  I., W\"urthwein, F., and {Open Sci Grid Executive Board} ({2011}).
\newblock {A Science Driven Production Cyberinfrastructure-the Open Science
  Grid}.
\newblock {\em {Journal of GRID Computing}}, {9}({2, Sp. Iss. SI}):{201--218}.

\bibitem[Applegate et~al., 2007]{10.5555/1374811}
Applegate, D.~L., Bixby, R.~E., Chvatal, V., and Cook, W.~J. (2007).
\newblock {\em The Traveling Salesman Problem: A Computational Study (Princeton
  Series in Applied Mathematics)}.
\newblock Princeton University Press, USA.

\bibitem[Asta and \"Ozcan, 2015]{Asta:2015}
Asta, S. and \"Ozcan, E. (2015).
\newblock A tensor-based selection hyper-heuristic for cross-domain heuristic
  search.
\newblock {\em Information Sciences}, 299:412 -- 432.

\bibitem[Asta et~al., 2016]{Asta:2016}
Asta, S., \"Ozcan, E., and Curtois, T. (2016).
\newblock A tensor based hyper-heuristic for nurse rostering.
\newblock {\em Knowledge-Based Systems}, 98:185 -- 199.

\bibitem[Bartz-Beielstein et~al., 2020]{Bartz2020benchmarking}
Bartz-Beielstein, T., Doerr, C., Bossek, J., Chandrasekaran, S., Eftimov, T.,
  Fischbach, A., Kerschke, P., Lopez-Ibanez, M., Malan, K.~M., Moore, J.~H.,
  Naujoks, B., Orzechowski, P., Volz, V., Wagner, M., and Weise, T. (2020).
\newblock Benchmarking in optimization: Best practice and open issues.
\newblock {\em arXiv preprint arXiv:2007.03488}.

\bibitem[Battiti and Tecchiolli, 1994]{battiti1994reactive}
Battiti, R. and Tecchiolli, G. (1994).
\newblock The reactive tabu search.
\newblock {\em ORSA journal on computing}, 6(2):126--140.

\bibitem[Bengio et~al., 2020]{BENGIO2020}
Bengio, Y., Lodi, A., and Prouvost, A. (2020).
\newblock Machine learning for combinatorial optimization: A methodological
  tour d’horizon.
\newblock {\em European Journal of Operational Research}.

\bibitem[Birattari et~al., 2002]{Birattari2002irace}
Birattari, M., St\"{u}tzle, T., Paquete, L., and Varrentrapp, K. (2002).
\newblock A racing algorithm for configuring metaheuristics.
\newblock In {\em Proceedings of the Genetic and Evolutionary Computation
  Conference}, GECCO '02, pages 11--18, San Francisco, CA, USA. Morgan Kaufmann
  Publishers Inc.

\bibitem[Bleuler et~al., 2003]{bltz2003a}
Bleuler, S., Laumanns, M., Thiele, L., and Zitzler, E. (2003).
\newblock {PISA---A Platform and Programming Language Independent Interface for
  Search Algorithms}.
\newblock In Fonseca, C.~M. et~al., editors, {\em Conference on Evolutionary
  Multi-Criterion Optimization {(EMO~2003)}}, volume 2632 of {\em LNCS}, pages
  494--508, Berlin. Springer.

\bibitem[Boussemart et~al., 2016]{DBLP:journals/corr/BoussemartLP16}
Boussemart, F., Lecoutre, C., and Piette, C. (2016).
\newblock {XCSP3:} an integrated format for benchmarking combinatorial
  constrained problems.
\newblock {\em CoRR}, abs/1611.03398.

\bibitem[Burke et~al., 2013]{Burke2013}
Burke, E.~K., Gendreau, M., Hyde, M., Kendall, G., Ochoa, G., {\"O}zcan, E.,
  and Qu, R. (2013).
\newblock Hyper-heuristics: a survey of the state of the art.
\newblock {\em Journal of the Operational Research Society}, 64(12):1695--1724.

\bibitem[Burke et~al., 2007]{burke:2007:graph}
Burke, E.~K., McCollum, B., Meisels, A., Petrovic, S., and Qu, R. (2007).
\newblock A graph-based hyper-heuristic for educational timetabling problems.
\newblock {\em European Journal of Operational Research}, 176(1):177--192.

\bibitem[Cahon et~al., 2004]{Cahon:2004:PFR:989094.989108}
Cahon, S., Melab, N., and Talbi, E.-G. (2004).
\newblock Paradis{EO}: A framework for the reusable design of parallel and
  distributed metaheuristics.
\newblock {\em Journal of {H}euristics}, 10(3).

\bibitem[Camacho{-}Villal{\'{o}}n et~al.,
  2019]{DBLP:journals/swarm/Camacho-Villalon19}
Camacho{-}Villal{\'{o}}n, C.~L., Dorigo, M., and St{\"{u}}tzle, T. (2019).
\newblock The intelligent water drops algorithm: why it cannot be considered a
  novel algorithm - {A} brief discussion on the use of metaphors in
  optimization.
\newblock {\em Swarm Intelligence}, 13(3-4):173--192.

\bibitem[Chakhlevitch and Cowling, 2008]{Chakhlevitch2008}
Chakhlevitch, K. and Cowling, P. (2008).
\newblock {\em Hyperheuristics: Recent Developments}, pages 3--29.
\newblock Springer Berlin Heidelberg, Berlin, Heidelberg.

\bibitem[Cloete et~al., 2008]{DBLP:conf/ijcnn/CloeteEP08}
Cloete, T., Engelbrecht, A.~P., and Pampar\`{a}, G. (2008).
\newblock Cilib: {A} collaborative framework for computational intelligence
  algorithms - part {II}.
\newblock In {\em Proceedings of the International Joint Conference on Neural
  Networks, {IJCNN} 2008, part of the {IEEE} World Congress on Computational
  Intelligence, {WCCI} 2008, Hong Kong, China, June 1-6, 2008}, pages
  1764--1773. {IEEE}.

\bibitem[Collberg et~al., 2015]{collberg2015repeatability}
Collberg, C., Proebsting, T., and Warren, A.~M. (2015).
\newblock Repeatability and benefaction in computer systems research.
\newblock Technical Report TR 14, University of Arizona.

\bibitem[Consoli et~al., 2014]{Consoli2014}
Consoli, P., Minku, L.~L., and Yao, X. (2014).
\newblock Dynamic selection of evolutionary algorithm operators based on online
  learning and fitness landscape metrics.
\newblock In {\em 10th International Conference on Simulated Evolution And
  Learning}, volume 8886 of {\em LNCS}. Springer.

\bibitem[Cowling et~al., 2001]{cowling:2001:HH}
Cowling, P., Kendall, G., and Soubeiga, E. (2001).
\newblock A hyperheuristic approach to scheduling a sales summit.
\newblock In Burke, E. and Erben, W., editors, {\em Practice and Theory of
  Automated Timetabling III}, volume 2079 of {\em Lecture Notes in Computer
  Science}, pages 176--190. Springer Berlin Heidelberg.

\bibitem[Cox et~al., 2001]{grid1}
Cox, S.~J., Fairman, M.~J., Xue, G., Wason, J.~L., and Keane, A.~J. (2001).
\newblock The grid: Computational and data resource sharing in engineering
  optimisation and design search.
\newblock In {\em 30th International Workshops on Parallel Processing (ICPP
  2001 Workshops), 3-7 September 2001, Valencia, Spain}, pages 207--212. IEEE
  Computer Society.

\bibitem[Dantzig, 1990]{Dantzig:1990:OSM}
Dantzig, G.~B. (1990).
\newblock A history of scientific computing.
\newblock chapter Origins of the Simplex Method, pages 141--151. ACM, New York,
  NY, USA.

\bibitem[De~Beukelaer et~al., 2017]{DeBeukelaer:2017}
De~Beukelaer, H., Davenport, G.~F., De~Meyer, G., and Fack, V. (2017).
\newblock James: An object-oriented java framework for discrete optimization
  using local search metaheuristics.
\newblock {\em Software: Practice and Experience}, 47(6):921--938.
\newblock spe.2459.

\bibitem[Di~Gaspero and Schaerf, 2003]{DiGaspero2003}
Di~Gaspero, L. and Schaerf, A. (2003).
\newblock Easylocal++: an object-oriented framework for the flexible design of
  local-search algorithms.
\newblock {\em Software: Practice and Experience}, 33(8):733--765.

\bibitem[Drake et~al., 2020]{drake:2019:survey}
Drake, J.~H., Kheiri, A., \"Ozcan, E., and Burke, E.~K. (2020).
\newblock Recent advances in selection hyper-heuristics.
\newblock {\em European Journal of Operational Research}, 285(2):405--428.

\bibitem[Durillo and Nebro, 2011]{Durillo2011760}
Durillo, J.~J. and Nebro, A.~J. (2011).
\newblock j{M}etal: A {J}ava framework for multi-objective optimization.
\newblock {\em Advances in Engineering Software}, 42(10).

\bibitem[Foster, 2005a]{Foster2005Globus}
Foster, I. ({2005}a).
\newblock {Globus Toolkit version 4: Software for service-oriented systems}.
\newblock In {Jin, H and Reed, D and Jiang, W}, editor, {\em {Network and
  Parallel Computing Proceedings}}, volume {3779} of {\em {Lecture Notes in
  Computer Science}}, pages {2--13}.

\bibitem[Foster, 2005b]{Foster2005Science}
Foster, I. (2005b).
\newblock Service-oriented science.
\newblock {\em Science}, 308(5723):814--817.

\bibitem[Fuellerer et~al., 2010]{Fuellerer2010}
Fuellerer, G., Doerner, K.~F., Hartl, R.~F., and Iori, M. (2010).
\newblock Metaheuristics for vehicle routing problems with three-dimensional
  loading constraints.
\newblock {\em European Journal of Operational Research}, 201(3):751 -- 759.

\bibitem[Garc{\'{\i}}a{-}Nieto et~al., 2007]{GarciaNieto07ROS}
Garc{\'{\i}}a{-}Nieto, J., Alba, E., and Chicano, J.~F. (2007).
\newblock Using metaheuristic algorithms remotely via {ROS}.
\newblock In {\em Proceedings of the 9th Annual Conference on Genetic and
  Evolutionary Computation}, GECCO '07, pages 1510--1510, New York, NY, USA.
  ACM.

\bibitem[Garc{\'{\i}}a{-}S{\'{a}}nchez et~al., 2013]{GarciaSanchez13SOEA}
Garc{\'{\i}}a{-}S{\'{a}}nchez, P., Gonz{\'{a}}lez, J., Castillo, P.~A., Arenas,
  M.~G., and Guerv{\'{o}}s, J. J.~M. (2013).
\newblock Service oriented evolutionary algorithms.
\newblock {\em Soft Comput.}, 17(6):1059--1075.

\bibitem[Glover and Laguna, 1997]{Glover:1997:TS:549765}
Glover, F. and Laguna, M. (1997).
\newblock {\em Tabu Search}.
\newblock Kluwer Academic Publishers, Norwell, MA, USA.

\bibitem[Goh et~al., 2017]{GOH201717}
Goh, S.~L., Kendall, G., and Sabar, N.~R. (2017).
\newblock Improved local search approaches to solve the post enrolment course
  timetabling problem.
\newblock {\em European Journal of Operational Research}, 261(1):17 -- 29.

\bibitem[Guerv{\'{o}}s et~al., 2020]{DBLP:conf/gecco/GuervosVG20}
Guerv{\'{o}}s, J. J.~M., Valdez, M.~G., and Galeano, S.~R. (2020).
\newblock Implementation matters, also in concurrent evolutionary algorithms.
\newblock In Coello, C. A.~C., editor, {\em {GECCO} '20: Genetic and
  Evolutionary Computation Conference, Companion Volume, Canc{\'{u}}n, Mexico,
  July 8-12, 2020}, pages 1591--1598. {ACM}.

\bibitem[Hackney et~al., 2006]{HACKNEY20061161}
Hackney, R., Xu, H., and Ranchhod, A. (2006).
\newblock Evaluating web services: Towards a framework for emergent contexts.
\newblock {\em European Journal of Operational Research}, 173(3):1161 -- 1174.

\bibitem[Hammond and Michaelson, 1999]{HammondMichaelson1999}
Hammond, K. and Michaelson, G. (1999).
\newblock {\em Research Directions in Parallel Functional Programming}.
\newblock Springer.

\bibitem[Hansen and Mladenovi{\'c}, 2001]{hansen2001variable}
Hansen, P. and Mladenovi{\'c}, N. (2001).
\newblock Variable neighborhood search: Principles and applications.
\newblock {\em European journal of operational research}, 130(3):449--467.

\bibitem[Hooker, 1995]{hooker:95}
Hooker, J. (1995).
\newblock Testing heuristics: We have it all wrong.
\newblock {\em Journal of Heuristics}, 1(1):33--42.

\bibitem[Hoos and St\"utzle, 2005]{HoosStutzle:04}
Hoos, H. and St\"utzle, T. (2005).
\newblock {\em Stochastic Local Search: Foundations \& Applications}.
\newblock Morgan Kaufmann.

\bibitem[Hughes, 1989]{Hughes}
Hughes, J. (1989).
\newblock Why functional programming matters.
\newblock {\em The Computer Journal}, 32(2):98--107.

\bibitem[Hunt and Thomas, 2001]{hunt2001art}
Hunt, A. and Thomas, D. (2001).
\newblock The art in computer programming.
\newblock {\em The Pragmatic Programmers, LLC}.

\bibitem[Imade et~al., 2004]{grid10}
Imade, H., Morishita, R., Ono, I., Ono, N., and Okamoto, M. (2004).
\newblock A grid-oriented genetic algorithm framework for bioinformatics.
\newblock {\em New Gen. Comput.}, 22(2):177--186.

\bibitem[Johnson, 2002]{Johnson2002}
Johnson, D.~S. (2002).
\newblock A theoretician’s guide to the experimental analysis of algorithms.
\newblock {\em Data structures, near neighbor searches, and methodology: fifth
  and sixth DIMACS implementation challenges}, 59:215--250.

\bibitem[Kendall et~al., 2016]{Kendall2016}
Kendall, G., Bai, R., B{\l}azewicz, J., De~Causmaecker, P., Gendreau, M., John,
  R., Li, J., McCollum, B., Pesch, E., Qu, R., et~al. (2016).
\newblock Good laboratory practice for optimization research.
\newblock {\em Journal of the Operational Research Society}, 67(4):676--689.

\bibitem[Khalloof et~al., 2018]{khalloof2018generic}
Khalloof, H., Jakob, W., Liu, J., Braun, E., Shahoud, S., Duepmeier, C., and
  Hagenmeyer, V. (2018).
\newblock A generic distributed microservices and container based framework for
  metaheuristic optimization.
\newblock In {\em Proceedings of the Genetic and Evolutionary Computation
  Conference Companion}, pages 1363--1370. ACM.

\bibitem[Kheiri, 2020]{kheiri:2019:irp}
Kheiri, A. (2020).
\newblock Heuristic sequence selection for inventory routing problem.
\newblock {\em Transportation Science}, 54(2):302--312.

\bibitem[Kheiri and {\"O}zcan, 2016]{kheiri:2016:iterated}
Kheiri, A. and {\"O}zcan, E. (2016).
\newblock An iterated multi-stage selection hyper-heuristic.
\newblock {\em European Journal of Operational Research}, 250(1):77--90.

\bibitem[Khichane et~al., 2008]{Khichane2008a}
Khichane, M., Albert, P., and Solnon, C. (2008).
\newblock {Integration of ACO in a constraint programming language}.
\newblock {\em Ant Colony Optimization and Swarm Intelligence}, pages 84--95.

\bibitem[Kirkpatrick et~al., 1983]{DBLP:journals/science/KirkpatrickGV83}
Kirkpatrick, S., Jr., D.~G., and Vecchi, M.~P. (1983).
\newblock Optimization by simmulated annealing.
\newblock {\em Science}, 220(4598):671--680.

\bibitem[Kocsis et~al., 2015]{haiku}
Kocsis, Z.~A., Brownlee, A. E.~I., Swan, J., and Senington, R. (2015).
\newblock Haiku - a {S}cala combinator toolkit for semi-automated composition
  of metaheuristics.
\newblock In Barros, M. and Labiche, Y., editors, {\em Search-Based Software
  Engineering}, volume 9275 of {\em Lecture Notes in Computer Science}, pages
  125--140. Springer International Publishing.

\bibitem[Kocsis and Swan, 2017]{DBLP:journals/corr/KocsisS17}
Kocsis, Z.~A. and Swan, J. (2017).
\newblock Dependency injection for programming by optimization.
\newblock {\em CoRR}, abs/1707.04016.

\bibitem[K{\"o}nig, 2020]{Konig2020.01.04.894873}
K{\"o}nig, M. (2020).
\newblock Executable simulation model of the liver.
\newblock {\em bioRxiv}.

\bibitem[Koza, 1992]{koza1992genetic}
Koza, J.~R. (1992).
\newblock {\em Genetic programming: on the programming of computers by means of
  natural selection}, volume~1.
\newblock MIT press.

\bibitem[Lim et~al., 2007]{grid8}
Lim, D., Ong, Y.-S., Jin, Y., Sendhoff, B., and Lee, B.-S. (2007).
\newblock Efficient hierarchical parallel genetic algorithms using grid
  computing.
\newblock {\em Future Generation Computer Systems}, 23(4):658 -- 670.

\bibitem[L\'{o}pez-Ib\'{a}\~{n}ez et~al.,
  2014]{Lopez-Ibanez:2014:TDS:2598394.2609846}
L\'{o}pez-Ib\'{a}\~{n}ez, M., Mascia, F., Marmion, M.-E., and St\"{u}tzle, T.
  (2014).
\newblock A template for designing single-solution hybrid metaheuristics.
\newblock In {\em GECCO Comp '14}, pages 1423--1426, New York, USA.

\bibitem[L{\'o}pez-Ib{\'a}{\~n}ez et~al., 2016]{LopDubStu2011irace}
L{\'o}pez-Ib{\'a}{\~n}ez, M., Dubois-Lacoste, J., {P\'erez C\'aceres}, L.,
  Birattari, M., and St{\"u}tzle, T. (2016).
\newblock The irace package: Iterated racing for automatic algorithm
  configuration.
\newblock {\em Operations Research Perspectives}, 3:43 -- 58.

\bibitem[Lukasiewycz et~al., 2011]{opt4jpaper}
Lukasiewycz, M., Gla{\ss}, M., Reimann, F., and Teich, J. (2011).
\newblock Opt4j - a modular framework for meta-heuristic optimization.
\newblock In {\em Proceedings of the Genetic and Evolutionary Computing
  Conference (GECCO 2011)}, pages 1723--1730, Dublin, Ireland.

\bibitem[Luke, 2010]{Luke:ECJ}
Luke, S. (2010).
\newblock The {ECJ} owner's manual.
\newblock \url{http://www.cs.gmu.edu/~eclab/projects/ecj}.

\bibitem[Luke, 2017]{Luke:2017}
Luke, S. (2017).
\newblock {ECJ} then and now.
\newblock In {\em Proceedings of the Genetic and Evolutionary Computation
  Conference Companion}, GECCO '17, pages 1223--1230, New York, NY, USA. ACM.

\bibitem[Malan and Engelbrecht, 2014]{Malan2014}
Malan, K.~M. and Engelbrecht, A.~P. (2014).
\newblock {\em Fitness Landscape Analysis for Metaheuristic Performance
  Prediction}, pages 103--132.
\newblock Springer Berlin Heidelberg, Berlin, Heidelberg.

\bibitem[Manna and Waldinger, 1980]{manna1980deductive}
Manna, Z. and Waldinger, R. (1980).
\newblock A deductive approach to program synthesis.
\newblock {\em ACM Transactions on Programming Languages and Systems (TOPLAS)},
  2(1):90--121.

\bibitem[Marmion et~al., 2013]{marmion2013automatic}
Marmion, M.-E., Mascia, F., L{\'o}pez-Ib{\'a}nez, M., and St{\"u}tzle, T.
  (2013).
\newblock Automatic design of hybrid stochastic local search algorithms.
\newblock In {\em International workshop on hybrid metaheuristics}, pages
  144--158. Springer.

\bibitem[Martin et~al., 2016]{Martin2016}
Martin, S., Ouelhadj, D., Beullens, P., Ozcan, E., Juan, A.~A., and Burke,
  E.~K. (2016).
\newblock A multi-agent based cooperative approach to scheduling and routing.
\newblock {\em European Journal of Operational Research}, 254(1):169 -- 178.

\bibitem[Merelo-Guerv{\'o}s et~al., 2003]{guervos2003specifying}
Merelo-Guerv{\'o}s, J.~J., Castillo-Valdivieso, P.~{\'A}., Romero-L{\'o}pez,
  G., and Garc{\'\i}a-Arenas, M. (2003).
\newblock Specifying evolutionary algorithms in xml.
\newblock In {\em International Work-Conference on Artificial Neural Networks},
  pages 502--509. Springer.

\bibitem[{Merelo Guerv{\'{o}}s} and Valdez, 2018]{Merelo18Stateless}
{Merelo Guerv{\'{o}}s}, J.~J. and Valdez, J. M.~G. (2018).
\newblock Mapping evolutionary algorithms to a reactive, stateless
  architecture: using a modern concurrent language.
\newblock In Aguirre, H.~E. and Takadama, K., editors, {\em Proceedings of the
  Genetic and Evolutionary Computation Conference Companion, {GECCO} 2018,
  Kyoto, Japan, July 15-19, 2018}, pages 1870--1877. {ACM}.

\bibitem[Miranda et~al., 2017]{MIRANDA2017340}
Miranda, P.~B., Prudêncio, R.~B., and Pappa, G.~L. (2017).
\newblock H3ad: A hybrid hyper-heuristic for algorithm design.
\newblock {\em Information Sciences}, 414(Supplement C):340 -- 354.

\bibitem[Munawar et~al., 2010]{gridUFO}
Munawar, A., Wahib, M., Munetomo, M., and Akama, K. (2010).
\newblock The design, usage, and performance of gridufo: A grid based unified
  framework for optimization.
\newblock {\em Future Generation Computer Systems}, 26(4):633 -- 644.

\bibitem[Nagata and Kobayashi, 1997]{DBLP:conf/icga/NagataK97}
Nagata, Y. and Kobayashi, S. (1997).
\newblock Edge assembly crossover: {A} high-power genetic algorithm for the
  travelling salesman problem.
\newblock In {\em Proceedings of the 7th International Conference on Genetic
  Algorithms, MI, USA}.

\bibitem[Nallaperuma et~al., 2014]{Nallaperuma2014}
Nallaperuma, S., Wagner, M., and Neumann, F. (2014).
\newblock {\em Parameter Prediction Based on Features of Evolved Instances for
  Ant Colony Optimization and the Traveling Salesperson Problem}, pages
  100--109.
\newblock Springer International Publishing, Cham.

\bibitem[Nallaperuma et~al., 2015]{Nallaperuma2015}
Nallaperuma, S., Wagner, M., and Neumann, F. (2015).
\newblock Analyzing the effects of instance features and algorithm parameters
  for max–min ant system and the traveling salesperson problem.
\newblock {\em Frontiers in Robotics and AI}, 2:18.

\bibitem[Neumann et~al., 2014]{Neumann:2014:EET:2598394.2609850}
Neumann, G., Swan, J., Harman, M., and Clark, J.~A. (2014).
\newblock The executable experimental template pattern for the systematic
  comparison of metaheuristics: Extended abstract.
\newblock In {\em Proceedings of the Companion Publication of the 2014 Annual
  Conference on Genetic and Evolutionary Computation}, GECCO Comp '14, pages
  1427--1430, New York, NY, USA. ACM.

\bibitem[Nikzad et~al., 2021]{Nikzad2021}
Nikzad, E., Bashiri, M., and Abbasi, B. (2021).
\newblock A matheuristic algorithm for stochastic home health care planning.
\newblock {\em European Journal of Operational Research}, 288(3):753 -- 774.

\bibitem[Pampar\`{a} and Engelbrecht, 2015]{Pamp1512:Generic}
Pampar\`{a}, G. and Engelbrecht, A. (2015).
\newblock Towards a generic computational intelligence library: Preventing
  insanity.
\newblock In {\em 2015 IEEE Symposium Series on Computational Intelligence:
  IEEE Workshop on Computational Intelligence Tools (2015 IEEE WCIT)}, Cape
  Town, South Africa.

\bibitem[Pampar\`{a} and Engelbrecht, 2019]{10.1145/3319619.3326845}
Pampar\`{a}, G. and Engelbrecht, A.~P. (2019).
\newblock Evolutionary and swarm-intelligence algorithms through monadic
  composition.
\newblock In {\em Proceedings of the Genetic and Evolutionary Computation
  Conference Companion}, GECCO '19, page 1382–1390, New York, NY, USA.
  Association for Computing Machinery.

\bibitem[Pampar\`{a} et~al., 2008]{DBLP:conf/ijcnn/PamparaEC08}
Pampar\`{a}, G., Engelbrecht, A.~P., and Cloete, T. (2008).
\newblock Cilib: {A} collaborative framework for computational intelligence
  algorithms - part {I}.
\newblock In {\em Proceedings of the International Joint Conference on Neural
  Networks, {IJCNN} 2008, part of the {IEEE} World Congress on Computational
  Intelligence, {WCCI} 2008, Hong Kong, China, June 1-6, 2008}, pages
  1750--1757. {IEEE}.

\bibitem[Pappa et~al., 2014]{pappa:2014:contrasting}
Pappa, G.~L., Ochoa, G., Hyde, M.~R., Freitas, A.~A., Woodward, J., and Swan,
  J. (2014).
\newblock Contrasting meta-learning and hyper-heuristic research: the role of
  evolutionary algorithms.
\newblock {\em Genetic Programming and Evolvable Machines}, 15(1):3--35.

\bibitem[Parejo, 2016]{Parejo15}
Parejo, J.~A. (2016).
\newblock {MOSES:} {A} metaheuristic optimization software ecosystem.
\newblock {\em {AI} Commun.}, 29(1):223--225.

\bibitem[Parejo et~al., 2003]{Parejo2003}
Parejo, J.~A., Racero, J., Guerrero, F., Kwok, T., and Smith, K.~A. (2003).
\newblock {\em FOM: A Framework for Metaheuristic Optimization}, pages
  886--895.
\newblock Springer Berlin Heidelberg, Berlin, Heidelberg.

\bibitem[Parejo et~al., 2012]{Parejo2012}
Parejo, J.~A., Ruiz-Cort{\'e}s, A., Lozano, S., and Fernandez, P. (2012).
\newblock Metaheuristic optimization frameworks: a survey and benchmarking.
\newblock {\em Soft Computing}, 16(3):527--561.

\bibitem[Parkes et~al., 2015]{parkes:2015:software}
Parkes, A.~J., {\"O}zcan, E., and Karapetyan, D. (2015).
\newblock A software interface for supporting the application of data science
  to optimisation.
\newblock In {\em International Conference on Learning and Intelligent
  Optimization}, pages 306--311. Springer.

\bibitem[{Peer} et~al., 2005]{1501612}
{Peer}, E.~S., {Engelbrecht}, A.~P., {Pampar\`{a}}, G., and {Masiye}, B.~S.
  (2005).
\newblock Ciclops: computational intelligence collaborative laboratory of
  pantological software.
\newblock In {\em Proceedings 2005 IEEE Swarm Intelligence Symposium, 2005. SIS
  2005.}, pages 130--137.

\bibitem[Pellerin et~al., 2020]{PELLERIN2020395}
Pellerin, R., Perrier, N., and Berthaut, F. (2020).
\newblock A survey of hybrid metaheuristics for the resource-constrained
  project scheduling problem.
\newblock {\em European Journal of Operational Research}, 280(2):395 -- 416.

\bibitem[Popper, 1963]{popperconjectures}
Popper, K. (1963).
\newblock {\em {Conjectures and Refutations: The Growth of Scientific
  Knowledge}}.
\newblock Routledge classics. Routledge.

\bibitem[Prud'homme et~al., 2016]{chocoSolver}
Prud'homme, C., Fages, J.-G., and Lorca, X. (2016).
\newblock {\em Choco Solver Documentation}.
\newblock TASC, INRIA Rennes, LINA CNRS UMR 6241, COSLING S.A.S.

\bibitem[Puchinger and Raidl, 2005]{Puchinger2005}
Puchinger, J. and Raidl, G. (2005).
\newblock {Combining metaheuristics and exact algorithms in combinatorial
  optimization: A survey and classification}.
\newblock In {\em Artificial Intelligence and Knowledge Engineering
  Applications: a Bioinspired Approach}, pages 113--124. Springer.

\bibitem[Qu et~al., 2009]{qu:2009:timetabling}
Qu, R., Burke, E.~K., and McCollum, B. (2009).
\newblock Adaptive automated construction of hybrid heuristics for exam
  timetabling and graph colouring problems.
\newblock {\em European Journal of Operational Research}, 198(2):392--404.

\bibitem[Raidl, 2015]{Raidl2015}
Raidl, G.~R. (2015).
\newblock Decomposition based hybrid metaheuristics.
\newblock {\em European Journal of Operational Research}, 244(1):66 -- 76.

\bibitem[Rice, 1976]{rice1976}
Rice, J.~R. (1976).
\newblock The algorithm selection problem.
\newblock In Rubinoff, M. and Yovits, M.~C., editors, {\em Advances in
  Computers}, volume~15, pages 65 -- 118. Elsevier.

\bibitem[Rosenberg et~al., 2010]{RosenbergMLMBD10}
Rosenberg, F., M{\"{u}}ller, M.~B., Leitner, P., Michlmayr, A., Bouguettaya,
  A., and Dustdar, S. (2010).
\newblock Metaheuristic optimization of large-scale qos-aware service
  compositions.
\newblock In {\em 2010 {IEEE} International Conference on Services Computing,
  {SCC} 2010, Miami, Florida, USA, July 5-10, 2010}, pages 97--104. {IEEE}
  Computer Society.

\bibitem[Ross, 2014]{ross:2014:survey}
Ross, P. (2014).
\newblock {\em Search Methodologies: Introductory Tutorials in Optimization and
  Decision Support Techniques}, chapter Hyper-heuristics, pages 611--638.
\newblock Springer US.

\bibitem[Rotem-Gal-Oz et~al., 2012]{rotem2012soa}
Rotem-Gal-Oz, A., Bruno, E., and Dahan, U. (2012).
\newblock {\em SOA patterns}.
\newblock Manning.

\bibitem[Scheibenpflug et~al., 2012]{Scheibenpflug:2012:OKB:2330784.2330806}
Scheibenpflug, A., Wagner, S., Pitzer, E., and Affenzeller, M. (2012).
\newblock {O}ptimization {K}nowledge {B}ase: An open database for algorithm and
  problem characteristics and optimization results.
\newblock In {\em GECCO '12}, NY, USA. ACM.

\bibitem[Senington and Duke, 2013]{DecomposingMetaheuristicOperations}
Senington, R. and Duke, D. (2013).
\newblock Decomposing metaheuristic operations.
\newblock In Hinze, R., editor, {\em Implementation and Application of
  Functional Languages}, {LNCS}. Springer Berlin Heidelberg.

\bibitem[Smith-Miles et~al., 2014]{SmithMiles201412}
Smith-Miles, K., Baatar, D., Wreford, B., and Lewis, R. (2014).
\newblock Towards objective measures of algorithm performance across instance
  space.
\newblock {\em Computers \& Operations Research}, 45:12--24.

\bibitem[Song et~al., 2003]{grid2}
Song, W., Keane, A., and Cox, S. (2003).
\newblock Cfd-based shape optimisation with grid-enabled design search
  toolkits.
\newblock In {\em UK e-Science All Hands Meeting 2003}, pages 619--627. EPSRC.

\bibitem[Song et~al., 2004]{grid3}
Song, W., Ong, Y.~S., Ng, H.~K., Keane, A., Cox, S., and Lee, B.~S. (2004).
\newblock A service-oriented approach for aerodynamic shape optimisation across
  institutional boundaries.
\newblock In {\em Control, Automation, Robotics and Vision Conference, 2004.
  ICARCV 2004 8th}, volume~3, pages 2274 -- 2279.

\bibitem[S{\"o}rensen, 2013]{Sor13}
S{\"o}rensen, K. (2013).
\newblock Metaheuristics---the metaphor exposed.
\newblock {\em International Transactions on Operational Research},
  22(1):3--18.

\bibitem[S{\"o}rensen and Glover, 2013]{SorensenGlover2013}
S{\"o}rensen, K. and Glover, F.~W. (2013).
\newblock {\em Metaheuristics}, pages 960--970.
\newblock Springer US, Boston, MA.

\bibitem[{S\"orensen, Kenneth and Arnold, Florian and Palhazi Cuervo, Daniel},
  2019]{doi:10.1111/itor.12443}
{S\"orensen, Kenneth and Arnold, Florian and Palhazi Cuervo, Daniel} (2019).
\newblock {A critical analysis of the ``improved Clarke and Wright savings
  algorithm''}.
\newblock {\em International Transactions in Operational Research},
  26(1):54--63.

\bibitem[Soria-Alcaraz et~al., 2017]{soria:2017:methodology}
Soria-Alcaraz, J.~A., Ochoa, G., Sotelo-Figeroa, M.~A., and Burke, E.~K.
  (2017).
\newblock A methodology for determining an effective subset of heuristics in
  selection hyper-heuristics.
\newblock {\em European Journal of Operational Research}, 260(3):972--983.

\bibitem[St{\"u}tzle and L{\'o}pez-Ib{\'a}{\~n}ez, 2019]{stutzle2019automated}
St{\"u}tzle, T. and L{\'o}pez-Ib{\'a}{\~n}ez, M. (2019).
\newblock Automated design of metaheuristic algorithms.
\newblock In {\em Handbook of metaheuristics}, pages 541--579. Springer.

\bibitem[Sutter, 2005]{sutter2005free}
Sutter, H. (2005).
\newblock The free lunch is over: A fundamental turn toward concurrency in
  software.
\newblock {\em Dr. {D}obbs {J}ournal}, 30(3).

\bibitem[Swan et~al., 2019]{AOCPECJ}
Swan, J., Adriaensen, S., Barwell, A.~D., Hammond, K., and White, D.~R. (2019).
\newblock Extending the `open-closed principle' to automated algorithm
  configuration.
\newblock {\em Evolutionary Computation}, 27(1):173--193.

\bibitem[Swan et~al., 2015]{micstandards2015}
Swan, J., Adriaensen, S., Bishr, M., Burke, E.~K., Clark, J.~A., Causmaecker,
  P.~D., Durillo, J., Hammond, K., Hart, E., Johnson, C.~G., Kocsis, Z.~A.,
  Kovitz, B., Krawiec, K., Martin, S., Merelo, J.~J., Minku, L.~L., \"Ozcan,
  E., Pappa, G.~L., Pesch, E., Garcia-S\`anchez, P., Schaerf, A., Sim, K.,
  Smith, J., St\"utzle, T., Vo\ss, S., Wagner, S., and Yao, X. (2015).
\newblock A research agenda for metaheuristic standardization.
\newblock In {\em Proceedings of the Eleventh {M}etaheuristics {I}nternational
  {C}onference (MIC), Agadir, Morocco}.

\bibitem[Swan et~al., 2018]{recharacterizationhh}
Swan, J., De~Causmaecker, P., Martin, S., and {\"O}zcan, E. (2018).
\newblock {\em A Re-characterization of Hyper-Heuristics}, pages 75--89.
\newblock Springer International Publishing, Cham.

\bibitem[Taillard, 2005]{Taillard2005}
Taillard, {\'E}.~D. (2005).
\newblock Tutorial : Few guidelines for analyzing methods.
\newblock In {\em Metaheuristic Interantional Conference (MIC'05) proceedings}.

\bibitem[Taylor, 2019]{TAYLOR20191}
Taylor, S.~J. (2019).
\newblock Distributed simulation: state-of-the-art and potential for
  operational research.
\newblock {\em European Journal of Operational Research}, 273(1):1 -- 19.

\bibitem[Thabtah and Cowling, 2008]{Thabtah:08}
Thabtah, F. and Cowling, P. (2008).
\newblock Mining the data from a hyperheuristic approach using associative
  classification.
\newblock {\em Expert Systems with Applications}, 34(2):1093--1101.

\bibitem[Valipour et~al., 2009]{Valipour09surveysoa}
Valipour, M.~H., Amirzafari, B., Maleki, K.~N., and Daneshpour, N. (2009).
\newblock A brief survey of software architecture concepts and service oriented
  architecture.
\newblock In {\em Computer Science and Information Technology, 2009. ICCSIT
  2009. 2nd IEEE International Conference on}, pages 34--38.

\bibitem[{Van Hentenryck} and Milano, 2011]{CPAIOR-tenyears}
{Van Hentenryck}, P. and Milano, M., editors (2011).
\newblock {\em Hybrid Optimization: The Ten Years of CPAIOR}, volume~45 of {\em
  Springer Optimization and Its Applications}.
\newblock Springer, Berlin, Germany.

\bibitem[Wagner and Affenzeller, 2005]{Wagner2005}
Wagner, S. and Affenzeller, M. (2005).
\newblock {\em Adaptive and Natural Computing Algorithms: Proceedings of the
  International Conference in Coimbra, Portugal, 2005}, chapter HeuristicLab: A
  Generic and Extensible Optimization Environment, pages 538--541.
\newblock Springer Vienna, Vienna.

\bibitem[Wagner et~al., 2014]{HEURISTICLAB}
Wagner, S., Kronberger, G., Beham, A., Kommenda, M., Scheibenpflug, A., Pitzer,
  E., Vonolfen, S., Kofler, M., Winkler, S., Dorfer, V., and Affenzeller, M.
  (2014).
\newblock {\em Advanced Methods and Applications in Computational
  Intelligence}, volume~6, chapter Architecture and Design of the HeuristicLab
  Optimization Environment, pages 197--261.
\newblock Springer.

\bibitem[Weyland, 2010]{Weyland:2010}
Weyland, D. (2010).
\newblock A rigorous analysis of the harmony search algorithm: How the research
  community can be misled by a "novel" methodology.
\newblock {\em Int. J. Appl. Metaheuristic Comput.}, 1(2):50--60.

\bibitem[Wolpert and Macready, 1997]{Wolpert:1997:NFL:2221336.2221408}
Wolpert, D.~H. and Macready, W.~G. (1997).
\newblock No free lunch theorems for optimization.
\newblock {\em Trans. Evol. Comp}, 1(1):67--82.

\bibitem[Woodward et~al., 2014]{Woodward:2014:CDP:2598394.2609848}
Woodward, J., Swan, J., and Martin, S. (2014).
\newblock {The `{C}omposite' {D}esign {P}attern in Metaheuristics}.
\newblock In {\em GECCO '14 Companion}, New York, USA.

\bibitem[Xu et~al., 2008]{xu2008satzilla}
Xu, L., Hutter, F., Hoos, H.~H., and Leyton-Brown, K. (2008).
\newblock {SATzilla: portfolio-based algorithm selection for SAT}.
\newblock {\em {Journal of Artificial Intelligence Research}}, 32:565--606.

\end{thebibliography}

\newpage
\appendix
\section{\added{The `Automated Open-Closed Principle'}\label{sec:automatedopenclosedprinciple}}
\msgtoreviewer{This content was originally part of Section \ref{automatedassembly}.}

\removed{The default practice in metaheuristic implementations is to make ad hoc use
of environmental state by accessing hidden non-local variables; in fact, all but
a few [55, 69] follow this kind of practice. This runs counter to automated assembly requirements: in order to present them to a configuration tool, such
variables must be manually gathered. An elementary alternative to this ad hoc
practice is to use manual state threading, as can be seen in Listing 2 where instead of defining the environmental state history as a shared or global variable
that any of the three components access and update directly, perturb, accept,
and finished each take an additional parameter that is used to thread the environmental state through the search algorithm. Manually threading the state
through the algorithm as in Listing 2 is error prone, since the framework implementer must ensure that the correct value of the state is passed to the correct
stage of the algorithm. It is instead desirable to use a mechanism that implicitly
performs state threading in a well-defined and consistent manner.}\added{
Existing metaheuristic implementations handle \textit{environmental state} in different ways. Many make \emph{ad hoc} use of non-local variables to share information between components. This runs counter to automated assembly requirements: in order to present them to a configuration tool, such dependencies must be specified manually, i.e. the configuration space cannot be derived from the implementation automatically. That being said, some implementations treat environmental state in a more principled manner (e.g.\ \cite{DiGaspero2003,Cahon:2004:PFR:989094.989108}). Here, metaheuristics are typically described as compositions of generically typed, stateful components having a well-defined interface controlling access to the encapsulated state. A framework then explicitly passes the shared state between subordinate components. While it has been demonstrated in prior-art that such implementations can be successfully coupled to configuration tools to perform bottom-up automated assembly \cite{marmion2013automatic, stutzle2019automated} \textit{in the small}, they do not support the open-ended extension we propose is required \textit{in the large} (as explained in Section~\ref{automatedassembly}).}

In software engineering, a framework which can be configured from an open-ended palette of components while remaining unchanged is said to conform to the ``Open-Closed principle'' (`a framework should be \emph{closed} to modification, but \emph{open} to extension by new components'). We have extended this principle \cite{AOCPECJ} to incorporate the behavior required to support automated design, yielding the ``Automated Open Closed Principle'' (AOCP); that paper discusses the issue in greater technical detail and describes experiments with a suitably equipped algorithm configurator. The adoption of the AOCP provides open-ended reuse of components, and thus a systematic approach to the automated exploration of the metaheuristic design space. A key aspect is that components must be ``pure functional'', as a first approximation\footnote{The interested reader is referred to the wealth of literature on `referential transparency'} this can be interpreted as meaning: 
\begin{itemize}
\item They do not rely on hidden state.
\item For the same argument, they always return the same result.
\end{itemize}
\noindent Note that aforementioned metaheuristic implementations using global variables and/or stateful components clearly violate the AOCP. An alternative to \emph{state encapsulation} is the use of \emph{state threading} \cite{haiku,Merelo18Stateless}, as can be seen in Listing~\ref{simplelshistory} where \lstinline|perturb|, \lstinline|accept|, and \lstinline|finished| each take an additional parameter that is used to \emph{thread} the environmental state through the search algorithm. More generally, using state threading, the desired signature for \lstinline$accept$ is of the form:
\[accept_{Sol,Env}: Sol \times Sol \times Env \rightarrow Sol \times Env\]
However, manually threading the state through the algorithm (as in Listing~\ref{simplelshistory}) of Section \ref{sec:approach} is error prone, since the framework implementer must ensure that the correct value of the state is passed to the correct stage of the algorithm. It is instead desirable to use a mechanism that \emph{implicitly} performs state threading in a \emph{well-defined} and \emph{consistent} manner.


In functional programming, the problem of state propagation is addressed via a well-known design pattern: the \emph{State monad}. For the purposes of this article, we can simply consider a monad to be a principled means of sequencing computations whilst abstracting over possible side-effects (in this case, state manipulation). 

\begin{lstlisting}[float=tb,mathescape=true,caption=Local Search in Scala with State monad,label=lst:monadlensls]
type Perturb[Env,Sol]    = Sol => State[Env,Sol]
type Accept[Env,Sol]     = (Sol,Sol) => State[Env,Sol]  
type IsFinished[Env,Sol] = Sol => State[Env,Boolean]

class LocalSearch[Env,Sol] {

  def apply[Sol](incumbent : Sol,
                 perturb : Perturb[Env,Sol],
                 accept : Accept[Env,Sol],
                 finished : IsFinished[Env,Sol]) : State[Env,Sol] = {
    def until(s : Sol) : State[Env,Sol] = {
      for {
        perturbed <- perturb(s)
        accepted <- accept(s, perturbed)
        c  <- finished(accepted)
        result <- if (c) {
                    State.pure[Env,Sol](accepted)
                  } else {
                    until(accepted)
                  }
      } yield result
    }
    
    for {
      result <- until(incumbent)
    } yield result 
  }
}

\end{lstlisting}

Functional languages such as Haskell and Scala provide syntactic sugar for monads. In particular, they allow monad operations to be chained together using syntax that looks like a traditional \emph{for} loop. This is illustrated in Listing \ref{lst:monadlensls}, a re-formulation of our local search example in Listing ~\ref{simplels} that uses the State monad. The state, in this case an integer representing the number of iterations, is implicitly threaded through each stage of the computation. This can be extended to yield a re-formulation of Listing~\ref{simplelshistory}, by defining that \lstinline|Env| allows access to  \lstinline|[Sol]|, the search trajectory.
Although the algorithm in Listing \ref{lst:monadlensls} looks similar to its imperative counterpart, the internal state is fully encapsulated within the definition of \verb=LocalSearch=. 

By virtue of open-ended support for state dependencies, the proposed approach therefore supports bottom-up automated assembly. Such an approach is less subject to human bias than the \emph{a priori}  prescription of a particular metaheuristic and therefore has relevance to foundational knowledge discovery efforts. In other areas of design (e.g.\ manufacturing), standardization has allowed a shift from the design of integrated systems to the design of individual components within the system. In metaheuristics, this reflects the natural trend for incorporating specialized problem- or solution- domain knowledge, i.e.\ a researcher can specialize in a particular kind of component such as acceptance criteria and determine their cross-domain ubiquity. 

It might be thought that a monadic workflow requires metaheuristic researchers to become expert functional programmers, so it should be emphasized that this workflow is a consequence of our proposed formulation, rather than a mandatory aspect. In particular, the intention is that core metaheuristic templates can be written monadically `once and for all', allowing non-expert users of these frameworks to obtain the benefits. Another minor but pleasing property is that the explicit denotation of state makes the parameter space of a component explicit, facilitating configuration via automated tools such as \emph{Irace} \cite{LopDubStu2011irace}.

This pure functional perspective also provides a number of other advantages~\cite{Hughes}, of particular relevance to large-scale and automated design of metaheuristics: they make it easy to reason about behavioural equivalence and coupling between components, hence improving transparency. Determinism and lack of side-effects yields reproducibility of behavior. Furthermore, a functional treatment of metaheuristics greatly facilitates architectures which can take advantage of abundant computing resources, e.g.\ thread-safe parallelism~\cite{HammondMichaelson1999} or `Service Oriented Architecture' (SOA) implementation via stateless web-services, as subsequently described in Section ~\ref{SOA}.

To our knowledge, the first proposed use of monads for state threading in metaheuristics was as part of the MitL initiative \cite{micstandards2015}, whilst the first concrete implementation subsequently appeared in CILib \cite{Pamp1512:Generic} and has since been further developed  \cite{10.1145/3319619.3326845}. 
Although there have been many frameworks and publications that describe `modular decompositions' of metaheuristics, to the best of our knowledge only MitL and CILib employ this principled approach to open-ended state dependencies. The  additional contribution of the MitL initiative in this respect is the use of the monadic approach to explicitly support automated assembly \cite{AOCPECJ}: of particular value in this respect is the fact that a strict type-system can be used to discriminate between stateless and stateful operations and to provide information about \emph{which} aspects of component behaviour contribute to solution quality, this being vital for the elimination of accidental complexity.

\section{\added{MitL Software Libraries}\label{sec:mitlsoftwarelibraries}}
\msgtoreviewer{This was originally Section \ref{sec:progress}.}

Realizing the MitL vision of community-level research based on shared scientific infrastructure requires the development of three central building blocks:

\begin{enumerate}
    \item Support for modular, extensible metaheuristic frameworks.
    \item Machine-readable descriptions of problems, heuristic components and results.
    \item A two-tier architecture defining both \emph{Programmatic} and a \emph{Service Oriented} interfaces, the latter being in direct correspondence with the former. 
\end{enumerate}

Taken together, these building blocks provide necessary support for the construction of a community knowledge base, in which fixed `reference versions' of metaheuristic templates can be configured with problems and components in an open-ended manner. Although the main purpose of this this paper is to describe the motivation and vision for MitL, the project has nonetheless made concrete implementation progress. The MitL repository (\url{https://github.com/MitLware}) contains various software libraries providing infrastructure support, together with a number of examples of how the proposed approach can be applied in practice. The infrastructure support libraries are:
\begin{itemize}
\item \texttt{MitLware-java}\\
This library contains Java interfaces for the `Metaheuristics in the Large' components (Perturb, Evaluate etc), as motivated by the discussion in Section \ref{automatedassembly}.
\item \texttt{mitl-support}\\
This library contains general utilities, metaheuristic-specific and otherwise. The former includes random selection and sampling.
\item \texttt{mitl-problem}\\
Example problem domains, defined in terms of the \texttt{MitLware-java} interfaces. The problem domains include: various bitvector problems, such as Checkerboard, Royal Road, Trap and HIFF; blocksworld; tower of Hanoi; the Iterated Prisoner's Dilemma; Magic Square; the $n$-puzzle; the De Jong suite of real-vector problems; SAT; the TSP; the Travelling-Thief Problem; Windfarm placement.
\item \texttt{mitl-solution}\\
Representations for ubiquitous candidate solution types (e.g.\ permutations, bit vectors and polynomials), as used in \texttt{mitl-problem}.
\end{itemize}

\noindent
Although the following examples happen to be implemented in Java/Scala, adoption at the Service Oriented Architecture level means that components written in other languages can nonetheless interoperate via standard serialization protocols (such as JSON or XML). For example, either or both of client or server in \texttt{mitl-soa-example} could be written in any language, as long as it is capable of serializing candidate solutions in JSON. The example applications include: 
\begin{itemize}
\item \texttt{mitl-whitebox-hyper-heuristics}\\
As an elementary example of the approach described in ``A Re-characterization of Hyper-Heuristics'', this demonstrates a whitebox analog of the hyflex hyper-heuristic framework which takes as input any problem domain (examples used are SAT, bin-packing, TSP, VRP) generically described via the XCSP constraint programming format. It then uses heuristic pattern matching to determine if the problem constraints are isomorphic to the TSP: if so, then the problem is rewritten on the fly to TSPLib format and a dedicated TSP solver (Concorde's `LINKERN' \cite{10.5555/1374811}) is used, if not then the generic Choco Solver \cite{chocoSolver} is invoked instead. 
\item \texttt{mitl-soa-example}\\
This provides a concrete demonstration of the `two tier architecture' described above: the MitL component interfaces defined in \texttt{MitLware-java} are `lifted' to the service level via RPC (Remote-Procedure Call) support. There is thus a 1-1 correspondence between local and remote component interfaces. A metaheuristic framework can therefore be transparently  configured with components that happen to be hosted remotely. A simple client-server example is provided, with remote invocation of a perturbation heuristic via json-RPC. The server-side implementation of perturb is actually achieved via a constraint solver, thereby giving another example of how one may freely mix between analytic    `OR-style' and empirical `metaheuristic-style' approaches.
\item \texttt{mitl-ecj-jmetal-interoperability-example}\\
The ECJ \cite{Luke:ECJ} and JMetal frameworks \cite{Durillo2011760} are both popular and widely used. However, it is not an easy task to achieve interoperability between them. This application shows how both can be represented as a MitL \texttt{Perturb} operator, allowing either to be interoperably invoked. 
\item \texttt{mitl-aocp}\\
This provides an example application of the our proposed \emph{Automated Open-Closed Principle} to automated algorithm configuration of the Traveling Salesperson Problem over a fixed algorithm framework. It uses ant-programming as a generative hyper-heuristic \cite{DBLP:journals/corr/KocsisS17} to automatically configure a local search framework with components which have different state dependencies. Further technical specifics of enabling communal research via extensible algorithm templates are described in detail in Swan et al. \cite{AOCPECJ}. 
\item \texttt{mitl-hyperion}\\
This provides extensible algorithm templates for several of the evolutionary algorithms described in `Essentials of Metaheuristics' \cite{Luke:ECJ}, as combinatorially instantiated in \texttt{mitl-aocp}, above.
\end{itemize}

\end{document}